\documentclass{article}

\usepackage[preprint]{./style/corl_2021} %

\usepackage[pdftex]{graphicx}
\graphicspath{{images-bin/}}
\graphicspath{{images-src/}}

\usepackage{xspace}
\usepackage{siunitx}[=v2]
\newcommand{\qty}{\SI}
\usepackage[textsize=scriptsize]{todonotes} 
\usepackage{url}
\usepackage{amsmath}
\usepackage{amssymb}
\usepackage[font=footnotesize]{caption}
\usepackage[font=footnotesize]{subcaption}
\usepackage{booktabs}
\usepackage{wrapfig}
\usepackage{algorithm}
\usepackage{comment}
\usepackage{algpseudocode}
\usepackage{cleveref}
\Crefname{equation}{Eq.}{Eqs.} %
\Crefname{figure}{Fig.}{Figs.}
\Crefname{section}{Sect.}{Sects.}

\usepackage{enumitem}

\definecolor{BrickRed}{cmyk}{0, .89, .5, 0}
\definecolor{Blue}{cmyk}{.8, .3, .2, 0}
\definecolor{Green}{cmyk}{1, 0.2, 1, 0}
\definecolor{Black}{cmyk}{1, 1, 1, 0}

\definecolor{aliceblue}{rgb}{0.94, 0.97, 1.0}
\definecolor{almond}{rgb}{0.94, 0.87, 0.8}
\definecolor{amber}{rgb}{1.0, 0.49, 0.0}
\definecolor{amber}{rgb}{1.0, 0.49, 0.0}

\definecolor{bleu}{HTML}{04D9E0}
\definecolor{oran}{HTML}{F5AC44}
\definecolor{vert}{HTML}{BED95F}

\DeclareRobustCommand{\TODOCRITICAL}[1]{
\todo[backgroundcolor=red!20,inline]{\textcolor{BrickRed}{\the\value{section}.\the\value{subsection}) #1}}
}

\DeclareRobustCommand{\sidenote}[1]{
\todo{\the\value{section}.\the\value{subsection}) #1}
}

\newcommand{\eg}{e.\,g.\ }
\newcommand{\ie}{i.\,e.\ }
\newcommand{\cf}{cf.}

\newcommand{\E}[2]{\operatorname{\mathbb{E}}_{#1}\left[#2\right]}
\newcommand{\Cont}{\operatorname{\mathcal{C}}}

\newcommand{\voidarg}{{\,\cdot\,}}

\newcommand{\sspace}{\mathcal{S}}
\newcommand{\aspace}{\mathcal{A}}
\newcommand{\state}{\mathbf{s}}
\newcommand{\reward}{r}
\newcommand{\action}{\mathbf{a}}
\newcommand{\at}{{\action_t}}
\newcommand{\st}{{\state_t}}

\newcommand{\policy}{\pi}
\newcommand{\entropy}{\mathcal{H}}
\newcommand{\pilatent}{\mathbf{z_\mu}}

\newcommand{\normal}{\mathcal{N}}
\newcommand{\ent}{\entropy}
\newcommand{\noise}{\epsilon}

\newcommand{\sde}{State-Dependent Exploration\xspace}
\newcommand{\SDE}{\textsc{SDE}\xspace}
\newcommand{\ourSDE}{\textit{g}\textsc{SDE}\xspace}
\newcommand{\aac}{\textsc{A2C}\xspace}
\newcommand{\ppo}{\textsc{PPO}\xspace}
\newcommand{\sac}{\textsc{SAC}\xspace}
\newcommand{\tddd}{\textsc{TD3}\xspace}

\newcommand{\hc}{\textsc{HalfCheetah}\xspace}
\newcommand{\hopper}{\textsc{Hopper}\xspace}
\newcommand{\ant}{\textsc{Ant}\xspace}
\newcommand{\walker}{\textsc{Walker2D}\xspace}

\title{Smooth Exploration for Robotic Reinforcement Learning}

\author{
  Antonin Raffin\\
  Robotics and Mechatronics Center (RMC)\\
  German Aerospace Center (DLR)
  Germany\\
  \texttt{antonin.raffin@dlr.de} \\
  \And
  Jens Kober\\
  Cognitive Robotics Department\\
  Delft University of Technology
  The Netherlands\\
  \texttt{j.kober@tudelft.nl} \\
  \And
  Freek Stulp\\
  Robotics and Mechatronics Center (RMC)\\
  German Aerospace Center (DLR)
  Germany\\
  \texttt{freek.stulp@dlr.de} \\
}

\begin{document}

\maketitle

\begin{abstract}

Reinforcement learning (RL) enables robots to learn skills from interactions with the real world.
In practice, the unstructured step-based exploration used in Deep RL -- often very successful in simulation -- leads to jerky motion patterns on real robots.
Consequences of the resulting shaky behavior are poor exploration, or even damage to the robot.
We address these issues by adapting state-dependent exploration (SDE)~\citep{ruckstiess2008state} to current Deep RL algorithms.
To enable this adaptation, we propose two extensions to the original SDE, using more general features and re-sampling the noise periodically, which leads to a new exploration method \textit{generalized state-dependent exploration} (\ourSDE).
We evaluate \ourSDE both in simulation, on PyBullet continuous control tasks, and directly on three different real robots: a tendon-driven elastic robot, a quadruped and an RC car.
The noise sampling interval of \ourSDE permits to have a compromise between performance and smoothness, which allows training directly on the real robots without loss of performance.
The code is available at \url{https://github.com/DLR-RM/stable-baselines3}.
\end{abstract}

\section{Introduction}
\label{sec:intro}

\begin{figure}[htb]
  \centering
  \includegraphics[width=0.49\columnwidth]{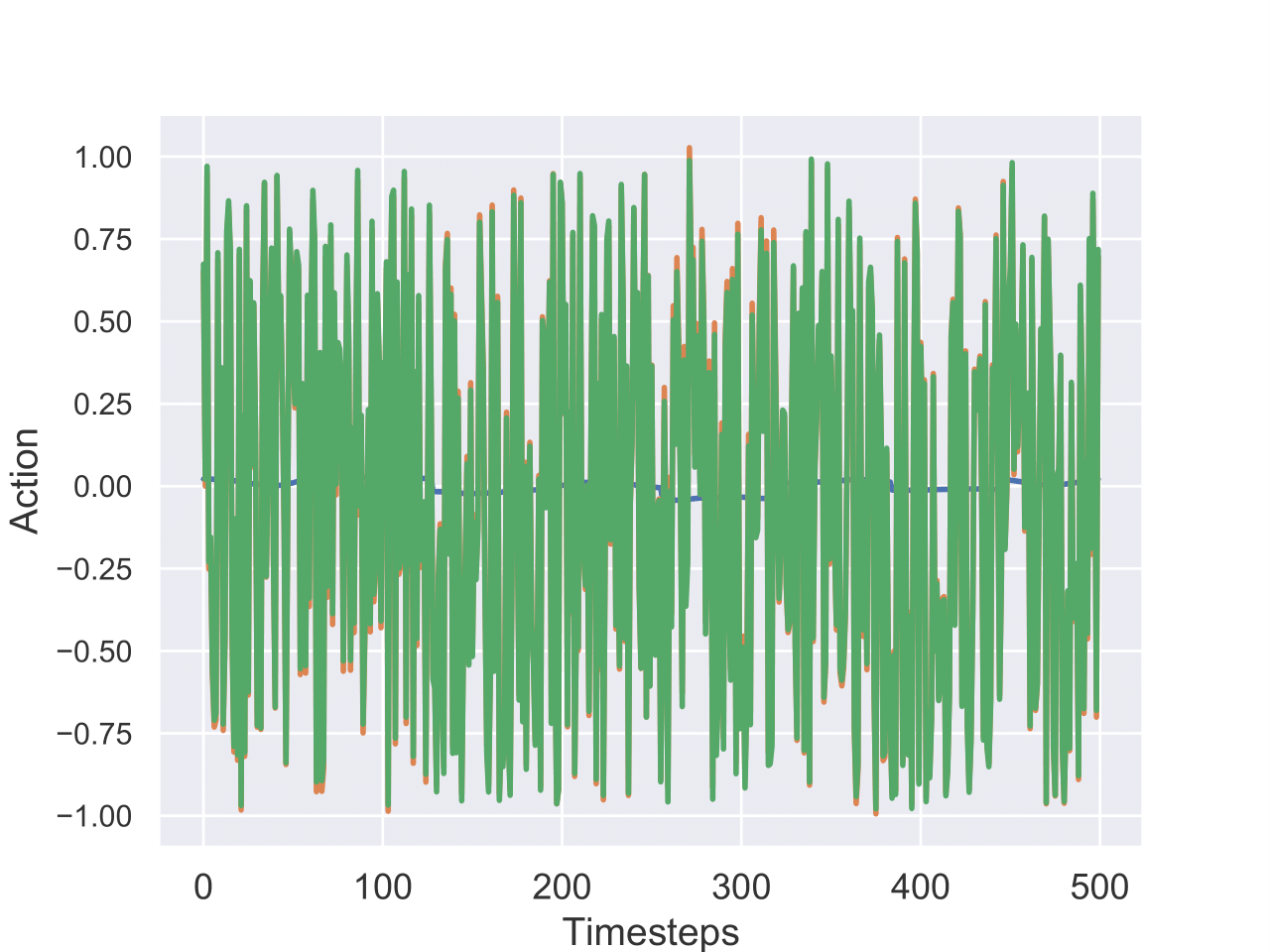}
  \hfill
  \includegraphics[width=0.49\columnwidth]{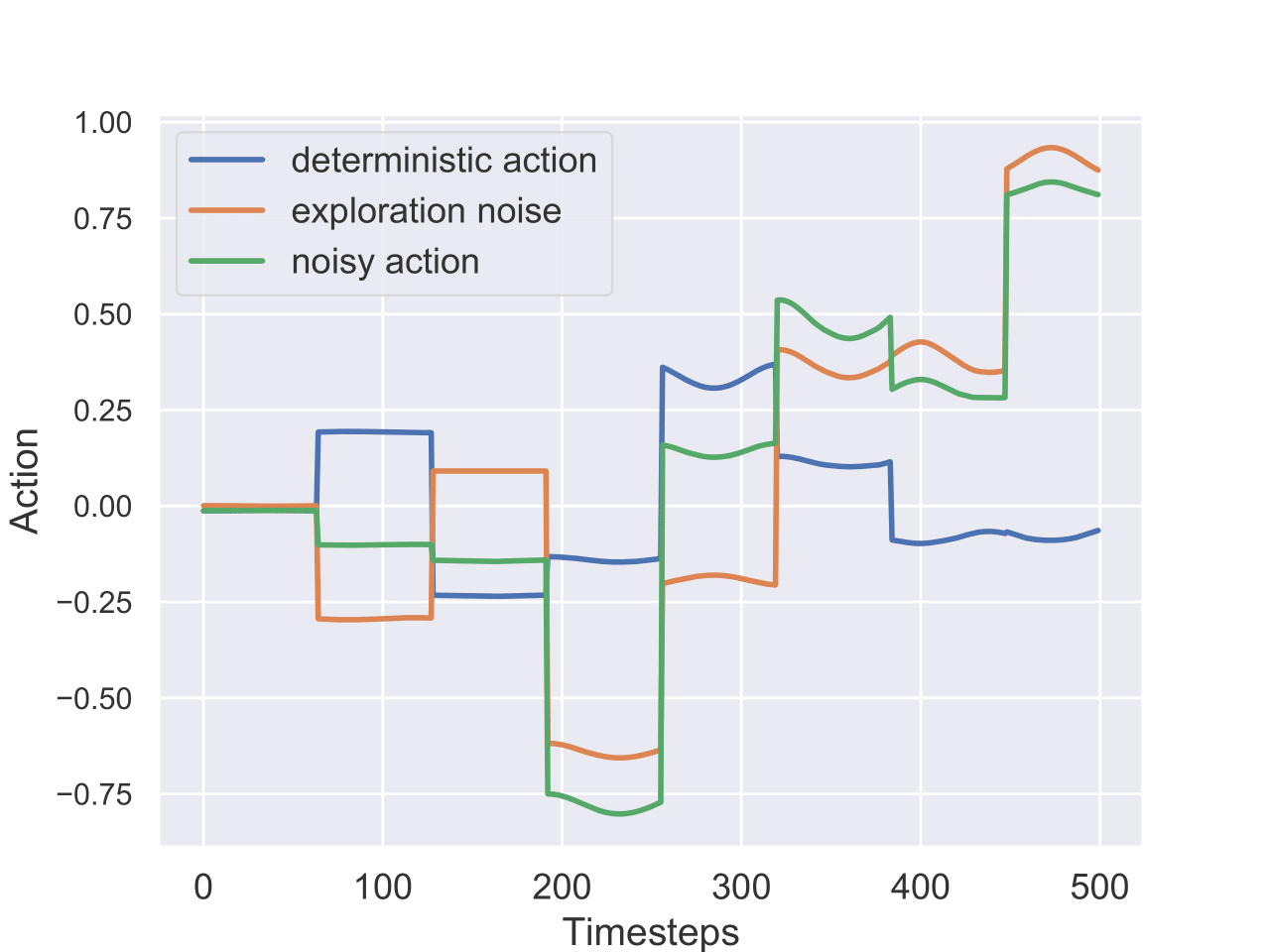}
  \caption{\label{fig:exploration_comparison} Left: unstructured exploration, as typically used in simulated RL. Right: \ourSDE provides smooth and consistent exploration.}
\end{figure}

One of the first robots that used artificial intelligence methods was called ``Shakey'', because it would shake a lot during operation~\citep{nilsson84shakey}.
Shaking has now again become quite prevalent in robotics, but for a different reason. When learning robotic skills with deep reinforcement learning (DeepRL), the de facto standard for exploration is to sample a noise vector $\noise_t$ from a Gaussian distribution independently at each time step $t$, and then adding it to the policy output.
This approach leads to the type of noise illustrated to the left in \Cref{fig:exploration_comparison}, and it can be very effective
in simulation~\citep{duan2016benchmarking, andrychowicz2018learning, fujimoto2018addressing, peng2018deepmimic, hwangbo2019learning}.

Unstructured exploration has also been applied to robotics~\citep{haarnoja2018applications, kendall2019learning}.
But for experiments on real robots, it has many drawbacks, which have been repeatedly pointed out~\citep{ruckstiess2008state, kober2009policy, ruckstiess2010exploring, stulp2013robot, deisenroth2013survey}:
1)~Sampling independently at each step leads to shaky behavior~\citep{mysore2021caps}, and noisy, jittery trajectories.
2)~The jerky motion patterns can damage the motors on a real robot, and lead to increased wear-and-tear.
3)~In the real world, the system acts as a low pass filter. Thus, consecutive perturbations may cancel each other, leading to poor exploration. This is particularly true for high control frequency~\citep{korenkevych2019autoregressive}.
4)~It causes a large variance which grows with the number of time-steps~\citep{kober2009policy, ruckstiess2010exploring, stulp2013robot}

In practice, we have observed all of these drawbacks on three real robots, including the tendon-driven robot David, depicted in \Cref{fig:david-neck}, which is the main experimental platform used in this work.
For all practical purposes, Deep RL with unstructured noise cannot be applied to David.

In robotics, multiple solutions have been proposed to counteract the inefficiency of unstructured noise.
These include correlated noise~\citep{haarnoja2018applications, korenkevych2019autoregressive}, low-pass filters~\citep{haarnoja2018learning, ha2020learning}, action repeat~\citep{neunert2020continuous} or lower level controllers~\citep{haarnoja2018learning, kendall2019learning}.
A more principled solution is to perform exploration in parameter space, rather than in action space~\citep{plappert2017parameter, pourchot2018cem}. This approach usually requires fundamental changes in the algorithm, and is harder to tune when the number of parameters is high.

\sde (\SDE)~\citep{ruckstiess2008state, ruckstiess2010exploring} was proposed as a compromise between exploring in parameter and action space. \SDE replaces the sampled noise with a state-dependent exploration function, which during an episode returns the same action for a given state. This results in smoother exploration and less variance per episode.

To the best of our knowledge, no Deep RL algorithm has yet been successfully combined with \SDE. We surmise that this is because the problem that it solves -- shaky, jerky movement -- is not as noticeable in simulation, which is the current focus of the community.

In this paper, we aim at reviving interest in \SDE as an effective method for addressing exploration issues that arise from using independently sampled Gaussian noise on real robots. Our concrete contributions, which also determine the structure of the paper, are:
\begin{enumerate}
  \item Highlighting the issues with unstructured Gaussian exploration (\Cref{sec:intro}).
  \item Adapting \SDE to recent Deep RL algorithms, and addressing some issues of the original formulation (\Cref{sec:sde,sec:gsde}).
  \item Evaluate the different approaches with respect to the compromise between smoothness and performance, and show the impact of the noise sampling interval (\Cref{sec:pybullet-envs,sec:ablation}).
  \item Successfully applying RL directly on three real robots: a tendon-driven robot, a quadruped and an RC car, without the need of a simulator or filters (\Cref{sec:exp-real}).
\end{enumerate}

\section{Background}
\label{sec:background}

In reinforcement learning, an agent interacts with its environment, usually modeled as a Markov Decision Process (MDP) $(\sspace, \aspace, p, \reward)$ where $\sspace$ is the state space, $\aspace$ the action space and $p(\state'|\state, \action)$ the transition function. At every step $t$, the agent performs an action $\action$ in state $\state$ following its policy $\policy : \sspace \mapsto \aspace$. It then receives a feedback signal in the next state $\state'$: the reward $\reward(\state, \action)$. The objective of the agent is to maximize the long-term reward. More formally, the goal is to maximize the expectation of the sum of discounted reward, over the trajectories $\rho_\policy$ generated using its policy~$\policy$:
\begin{align}
  \label{eq:rl_obj}
  \sum_t \E{(\st,\at)\sim\rho_\policy}{\gamma^t \reward(\st,\at)}
\end{align}
where $\gamma \in [0,1)$ is the discount factor and represents a trade-off between maximizing short-term and long-term rewards.
The agent-environment interactions are often broken down into sequences called \textit{episodes}, that end when the agent reaches a terminal state.

\subsection{Exploration in Action or Policy Parameter Space}

In the case of continuous actions, the exploration is commonly done in the \textit{action space}~\citep{schulman2015trust, lillicrap2015continuous, mnih2016asynchronous, schulman2017proximal, haarnoja2017reinforcement, fujimoto2018addressing}. At each time-step, a noise vector $\noise_t$ is independently sampled from a Gaussian distribution and then added to the controller output:
\begin{align}
  \label{eq:exploration_action_space}
  \at &= \mu(\st; \theta_\mu) + \noise_t, & \noise_t \sim \normal(0, \sigma^2)
\end{align}
where $\mu(\st)$ is the deterministic policy and $\policy(\at | \st) \sim \normal(\mu(\st), \sigma^2)$ is the resulting stochastic policy, used for exploration. $\theta_\mu$ denotes the parameters of the deterministic policy.\\
For simplicity, throughout the paper, we will only consider Gaussian distributions with diagonal covariance matrices. Hence, here, $\sigma$ is a vector with the same dimension as the action space $\aspace$.

Alternatively, the exploration can also be done in the \textit{parameter space}~\citep{ruckstiess2010exploring, plappert2017parameter, pourchot2018cem}:
\begin{align}
  \label{eq:exploration_parameter_space}
  \at &= \mu(\st; \theta_\mu + \noise), & \noise \sim \normal(0, \sigma^2)
\end{align}
at the beginning of an episode, the perturbation $\noise$ is sampled and added to the policy parameters $\theta_\mu$.
This usually results in more consistent exploration but becomes challenging with an increasing number of parameters~\citep{plappert2017parameter}.

\subsection{State-Dependent Exploration}
\label{sec:sde}

\textit{\sde(\SDE)}~\citep{ruckstiess2008state, ruckstiess2010exploring} is an intermediate solution that consists in adding noise as a function of the state $\st$, to the deterministic action $\mu(\st)$. At the beginning of an episode, the parameters $\theta_\noise$ of that exploration function are drawn from a Gaussian distribution. The resulting action $\at$ is as follows:
\begin{align}
  \label{eq:sde}
  \at &= \mu(\st; \theta_\mu) + \noise(\st; \theta_\noise), & \theta_\noise \sim \normal(0, \sigma^2)
\end{align}
This episode-based exploration is smoother and more consistent than the unstructured step-based exploration. Thus, during one episode, instead of oscillating around a mean value, the action $\action$ for a given state $\state$ will be the same.

\SDE should not be confused with unstructured noise where the variance can be state-dependent but the noise is still sampled at every step, as it is the case for \sac.

In the remainder of this paper, to avoid overloading notation, we drop the time subscript $t$, \ie we now write $\state$ instead of $\st$. $\state_j$ or $\action_j$ now refer to an element of the state or action vector.

In the case of a linear exploration function $\noise(\state; \theta_\noise) = \theta_\noise \state$, by operation on Gaussian distributions, \citet{ruckstiess2008state} show that the action element $\action_j$ is normally distributed:
\begin{align}
  \label{eq:sde-distri}
  \policy_j(\action_j|\state) \sim \normal(\mu_j(\state), \hat{\sigma_j}^2)
\end{align}
where $\hat{\sigma}$ is a diagonal matrix with elements $\hat{\sigma_j} = \sqrt{\sum_i{(\sigma_{ij} \state_i)^2}}$.

Because we know the policy distribution, we can obtain the derivative of the log-likelihood $\log \policy(\action|\state)$ with respect to the variance $\sigma$:
\begin{align}
  \label{eq:sde-logprob}
  \frac{\partial \log \policy(\action|\state)}{\partial \sigma_{ij}}
  &= \frac{(\action_j - \mu_j)^2 - \hat{\sigma_j}^2}{\hat{\sigma_j}^3}
  \frac{\state_i^2 \sigma_{ij}}{\hat{\sigma_{j}}}
\end{align}
This can be easily plugged into the likelihood ratio gradient estimator~\citep{williams1992simple}, which allows to adapt $\sigma$ during training.
\SDE is therefore compatible with standard policy gradient methods, while addressing most shortcomings of the unstructured exploration.

For a non-linear exploration function, the resulting distribution $\policy(\action|\state)$ is most of the time unknown. Thus, computing the exact derivative w.r.t.\ the variance is not trivial and may require approximate inference. As we focus on simplicity, we leave this extension for future work.

\section{Generalized \sde}
\label{sec:gsde}

Considering~\Cref{eq:sde-distri,eq:sde-logprob}, some limitations of the original formulation are apparent:
\begin{enumerate}[label=\roman*]
  \item The noise does not change during one episode, which is problematic if the episode length is long, because the exploration will be limited~\citep{hoof2017generalized}. \label{item:episode-sampling}
  \item The variance of the policy $\hat{\sigma_j} = \sqrt{\sum_i{(\sigma_{ij} \state_i)^2}}$ depends on the state space dimension (it grows with it), which means that the initial $\sigma$ must be tuned for each problem. \label{item:variance-task}
  \item There is only a linear dependency between the state and the exploration noise, which limits the possibilities. \label{item:linear-noise}
  \item The state must be normalized, as the gradient and the noise magnitude depend on the state magnitude. \label{item:gradient-issue}
\end{enumerate}

To mitigate the mentioned issues and adapt it to Deep RL algorithms, we propose two improvements:
\begin{enumerate}
  \item We sample the parameters $\theta_\noise$ of the exploration function every $n$ steps instead of every episode. \label{item:n-steps}
  \item Instead of the state $\state$, we can in fact use any features. We chose policy features $\pilatent(\state; \theta_\pilatent)$ (last layer before the deterministic output $\mu(\state) = \theta_\mu \pilatent(\state; \theta_\pilatent)$) as input to the noise function $\noise(\state; \theta_\noise) = \theta_\noise \pilatent(\state)$. \label{item:features}
\end{enumerate}

Sampling the parameters $\theta_\noise$ every $n$ steps tackles the issue~\ref{item:episode-sampling}. and yields a unifying framework~\citep{hoof2017generalized} which encompasses both unstructured exploration ($n=1$) and original \SDE ($n = \mathrm{episode\_length}$).
Although this formulation follows the description of Deep RL algorithms that update their parameters every $m$ steps, the influence of this crucial parameter on smoothness and performance was until now overlooked.

Using \textit{policy features} allows mitigating issues~\ref{item:variance-task}, \ref{item:linear-noise} and \ref{item:gradient-issue}: the relationship between the state $\state$ and the noise $\noise$ is non-linear and the variance of the policy only depends on the network architecture, allowing for instance to use images as input.
This formulation is therefore more general and includes the original \SDE description, when using state as input to the noise function or when the policy is linear.

We call the resulting approach \textit{generalized} \sde (\ourSDE).

\paragraph{Deep RL algorithms}
Integrating this updated version of \SDE into recent Deep RL algorithms, such as those listed in the appendix,
is straightforward. For those that rely on a probability distribution, such as \sac or \ppo, we can replace the original Gaussian distribution by the one from~\Cref{eq:sde-distri}, where the analytical form of the log-likelihood is known (\cf~\Cref{eq:sde-logprob}).

\section{Experiments}
\label{sec:experiments}

In this section, we study \ourSDE to answer the following questions:
\begin{itemize}
  \item How does \ourSDE compares to the original \SDE? What is the impact of each proposed modification?
  \item How does \ourSDE compares to other type of exploration noise in terms of compromise between smoothness and performance?
  \item How does \ourSDE performs on a real system?
\end{itemize}

\subsection{Compromise Between Smoothness and Performance}
\label{sec:pybullet-envs}

\paragraph{Experiment setup}

In order to compare \ourSDE to other type of exploration in terms of compromise between performance and smoothness,
we chose 4 locomotion tasks from the PyBullet~\citep{coumans2019bullet} environments: \hc, \ant, \hopper and \walker.
They are similar to the one found in OpenAI Gym~\citep{brockman2016openai} but the simulator is open source and they are harder to solve\footnote{\url{https://frama.link/PyBullet-harder-than-MuJoCo}}.
In this section, we focus on the \sac algorithm as it will be the one used on the real robot, although we report results for additional algorithms such as \ppo in the appendix.

To evaluate smoothness, we define a continuity cost $\Cont = 100 \times \E{t}{(\frac{\action_{t+1} - \at}{\Delta^{\action}_{\max}})^2}$ which yields values between 0 (constant output) and 100 (action jumping from one limit to another at every step). The continuity cost of the training $\Cont_\mathrm{train}$ is a proxy for the wear-and-tear of the robot.

We compare the performance of the following configurations: (a) no exploration noise, (b) unstructured Gaussian noise (original \sac implementation), (c) correlated noise (Ornstein–Uhlenbeck process~\citep{uhlenbeck1930theory} with $\sigma {=} 0.2$, OU noise in the figure), (d) adaptive parameter noise~\citep{plappert2017parameter} ($\sigma {=} 0.2$), (e) \ourSDE.
To decorrelate the exploration noise from the one due to parameter update, and to be closer to a real robot setting, we apply the gradient updates only at the end of each episode.

We fix the budget to 1 million steps and report the average score over 10 runs together with the average continuity cost during training and their standard error.
For each run, we test the learned policy on 20 evaluation episodes every 10000 steps, using the deterministic controller $\mu(\st)$.
Regarding the implementation,
we use a modified version of Stable-Baselines3~\citep{raffin2019baselines3} together with the RL Zoo training framework~\citep{raffin2020zoo3}.
The methodology we follow to tune the hyperparameters and their details can be found in the appendix.
The code we used to run the experiments and tune the hyperparameters can be found in the supplementary material.

\paragraph{Results}

\begin{wrapfigure}[18]{r}{0.5\textwidth}
    \vspace{-2em}
    \includegraphics[width=0.5\textwidth]{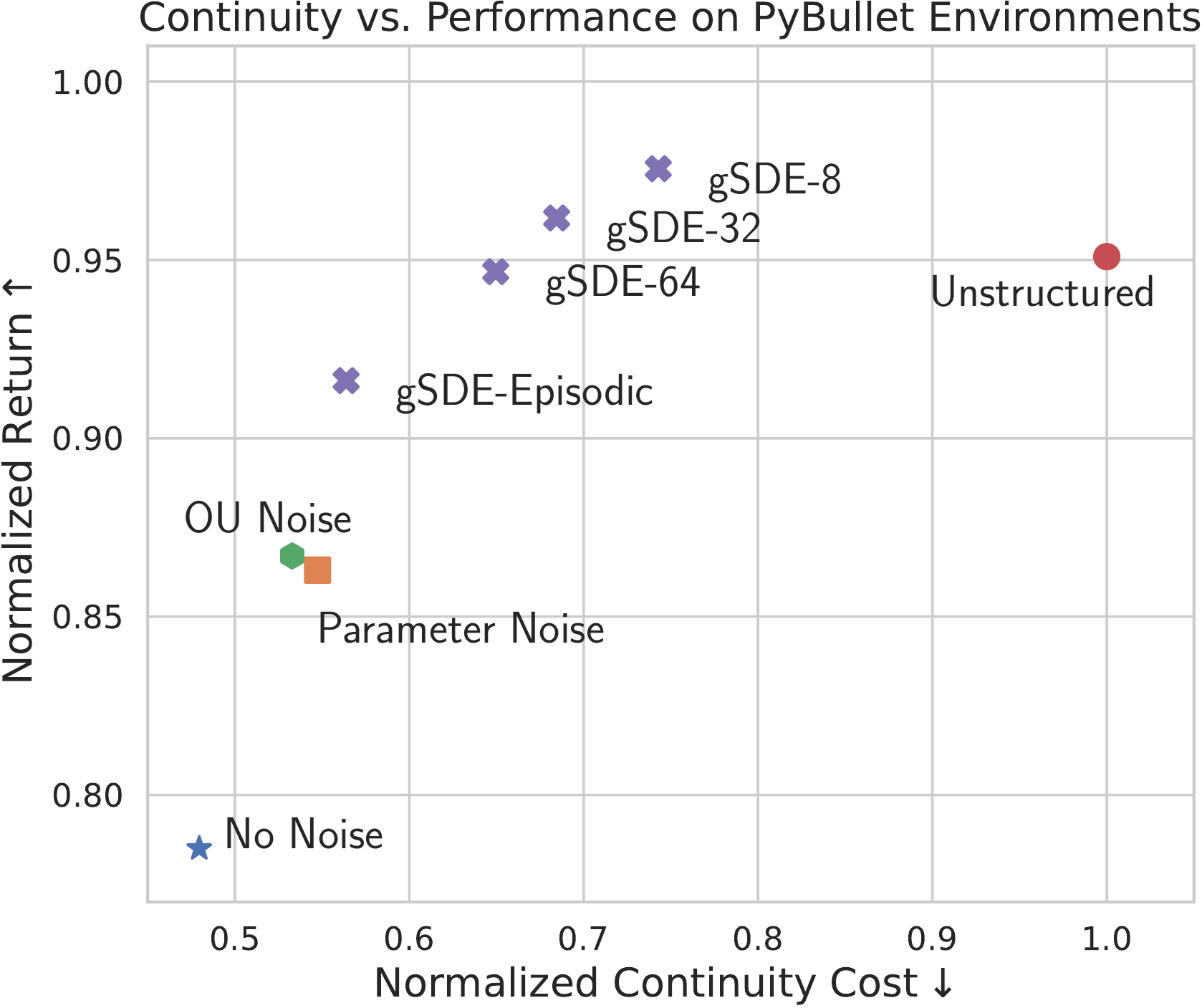}
    \caption{Normalized return and continuity cost of \sac on 4 PyBullet tasks with different type of exploration. \ourSDE provides a compromise between performance and smoothness.}
    \label{fig:pybullet-pareto}
\end{wrapfigure}

\begin{table}[h]
    \resizebox{\linewidth}{!}{
    \centering
        \begin{tabular}{l|ll|ll|ll|ll}
        \toprule
        \multicolumn{1}{l}{Algorithm} & \multicolumn{2}{c}{\hc} & \multicolumn{2}{c}{\ant} & \multicolumn{2}{c}{\hopper} & \multicolumn{2}{c}{\walker} \\
        \toprule
        \sac & Return $\uparrow$ & $\Cont_\mathrm{train} \downarrow$ & Return $\uparrow$ & $\Cont_\mathrm{train}\downarrow$ & Return $\uparrow$ & $\Cont_\mathrm{train}\downarrow$ & Return $\uparrow$ & $\Cont_\mathrm{train}\downarrow$ \\
        \midrule
        w/o noise & 2562 +/- 102  & \textbf{2.6}  +/- 0.1 & 2600 +/- 364 & \textbf{2.0} +/- 0.2  & 1661 +/- 270 & \textbf{1.8} +/- 0.1  & 2216 +/- 40 & \textbf{1.8} +/- 0.1 \\
        w/ unstructured & \textbf{2994} +/- 89 & 4.8  +/- 0.2 & \textbf{3394} +/- 64 & 5.1 +/- 0.1  & \textbf{2434} +/- 190 & 3.6 +/- 0.1  & 2225 +/- 35 & 3.6 +/- 0.1 \\
        w/ OU noise   & 2692 +/- 68 & 2.9  +/- 0.1 & 2849 +/- 267 & 2.3 +/- 0.0 & 2200 +/- 53 & 2.1 +/- 0.1  & 2089 +/- 25 & 2.0 +/- 0.0 \\
        w/ param noise   & 2834 +/- 54 &  2.9  +/- 0.1 & 3294 +/- 55 & \textbf{2.1} +/- 0.1 & 1685 +/- 279 & 2.2 +/- 0.1  & \textbf{2294} +/- 40 & \textbf{1.8} +/- 0.1\\
        \midrule
        \midrule
        w/ \ourSDE-8           & \textbf{2850} +/- 73 & 4.1 +/- 0.2 & \textbf{3459} +/- 52  & 3.9 +/- 0.2  & \textbf{2646} +/- 45 & 2.4 +/- 0.1  & \textbf{2341} +/- 45 & 2.5 +/- 0.1\\
        w/ \ourSDE-64          & \textbf{2970} +/- 132 & 3.5 +/- 0.1 & 3160 +/- 184 & 3.5 +/- 0.1  & 2476 +/- 99 & \textbf{2.0} +/- 0.1  & \textbf{2324} +/- 39 & 2.3 +/- 0.1 \\
        w/ \ourSDE-episodic  & 2741 +/- 115 & 3.1 +/- 0.2 & 3044 +/- 106 & 2.6 +/- 0.1  & 2503 +/- 80 & \textbf{1.8} +/- 0.1  & \textbf{2267} +/- 34 & 2.2 +/- 0.1 \\
        \bottomrule
        \multicolumn{9}{c}{}
        \end{tabular}
    }

  \caption{
  Detailed return and continuity cost results for \sac with different type of exploration on PyBullet environments. We report the mean and standard error over 10 runs of 1 million steps. For each benchmark, we highlight the results of the method(s) with the best mean when the difference is statistically significant.
  }
  \label{tab:results-bullet}
\end{table}

\Cref{tab:results-bullet,fig:pybullet-pareto} shows the results on the PyBullet tasks and the compromise between continuity and performance.
Without any noise (``No Noise'' in the figure), \sac is still able to solve partially those tasks thanks to a shaped reward, but it has the highest variance in the results.
Although the correlated and parameter noise yield lower continuity cost during training, it comes at a cost of performance.
\ourSDE is able to achieve a good compromise between unstructured exploration and correlated noise by making use of the noise repeat parameter.
\ourSDE-8 (sampling the noise every 8 steps) even achieves better performance with a lower continuity cost at train time.
Such behavior is what is desirable for training on a real robot: we must minimize wear-and-tear at training while still obtaining good performance at test time.

\subsection{Comparison to the Original \SDE}
\label{sec:ablation}

In this section, we investigate the contribution of the proposed modifications to the original \SDE: sampling the exploration function parameters every $n$ steps and using policy features as input to the noise function.

\begin{figure}[h]
  \begin{minipage}[t]{.45\linewidth}
    \centering\includegraphics[width=\linewidth]{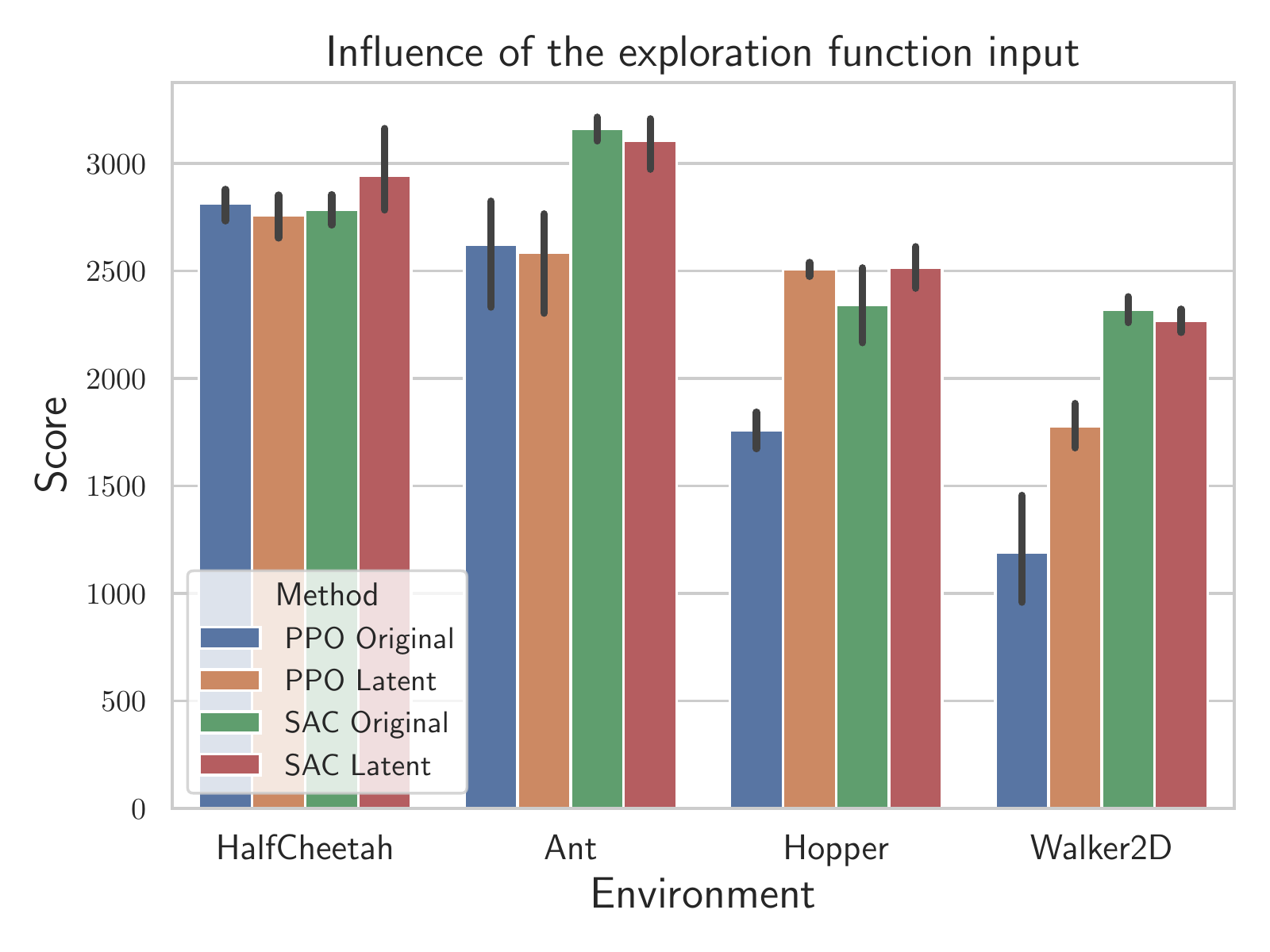}
    \subcaption{Exploration function input}
    \label{fig:sde-features}
  \end{minipage}
  \hfill
  \begin{minipage}[t]{.45\linewidth}
    \centering\includegraphics[width=\linewidth]{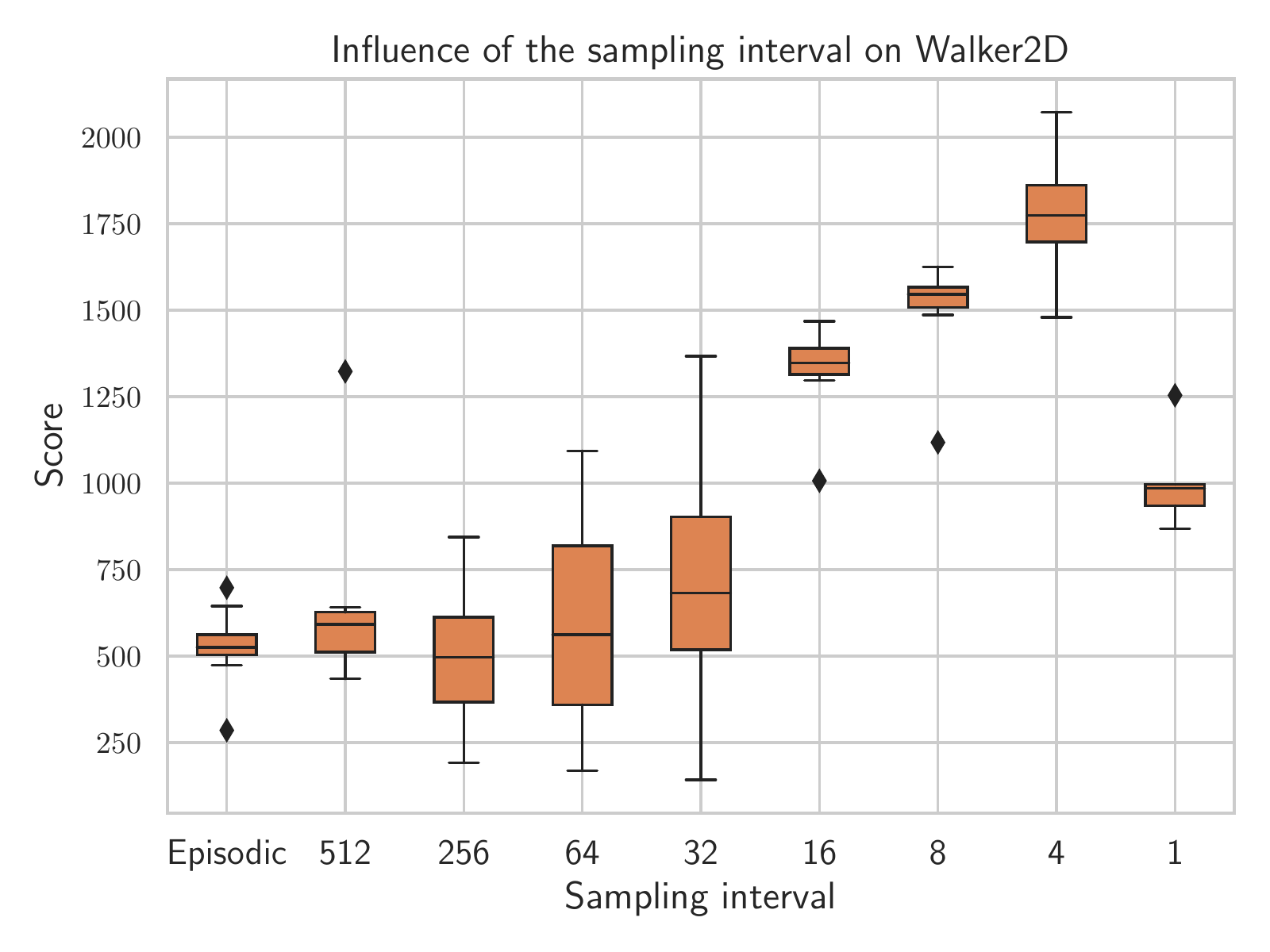}
    \subcaption{Sampling interval (\ppo on \walker)}
    \label{fig:ppo-sample-freq}
  \end{minipage}
  \caption{Impact of \ourSDE modifications over the original \SDE on PyBullet tasks. (a) Influence of the input to the exploration function $\noise(\state; \theta_\noise)$ for \sac and \ppo: using latent features from the policy $\pilatent$ (Latent) is usually better than using the state $\state$ (Original). (b) The frequency of sampling the noise function parameters is crucial for \ppo with \ourSDE.}
\end{figure}

\paragraph{Sampling Interval}

\ourSDE is a $n$-step version of \SDE, where $n$ allows interpolating between the unstructured exploration $n=1$ and the original \SDE per-episode formulation.
This interpolation allows to have a compromise between performance and smoothness at train time (\cf\ \Cref{tab:results-bullet,fig:pybullet-pareto}).
\Cref{fig:ppo-sample-freq} shows the importance of that parameter for \ppo on the \walker task.
If the sampling interval is too large, the agent would not explore enough during long episodes.
On the other hand, with a high sampling frequency $n \approx 1$, the issues mentioned in~\Cref{sec:intro} arise.

\paragraph{Policy features as input}
~\Cref{fig:sde-features} shows the effect of changing the exploration function input for \sac and \ppo. Although it varies from task to task, using policy features (``latent'' in the figure) is usually beneficial, especially for \ppo. It also requires less tuning and no normalization as it depends only on the policy network architecture. Here, the PyBullet tasks are low dimensional and the state space size is of the same order, so no careful per-task tuning is needed. Relying on features also allows learning directly from pixels, which is not possible in the original formulation.

Compared to the original \SDE, the two proposed modifications are beneficial to the performance, with the noise sampling interval $n$ having the most impact.
Fortunately, as shown in~\Cref{tab:results-bullet,fig:pybullet-pareto}, it can be chosen quite freely for \sac.
In the appendix, we provide an additional ablation study that shows \ourSDE is robust to the choice of the initial exploration variance.

\subsection{Learning to Control a Tendon-Driven Elastic Robot}
\label{sec:exp-real}
\begin{figure}[h]
  \begin{minipage}[t]{.5\linewidth}
    \centering\includegraphics[width=0.8\linewidth]{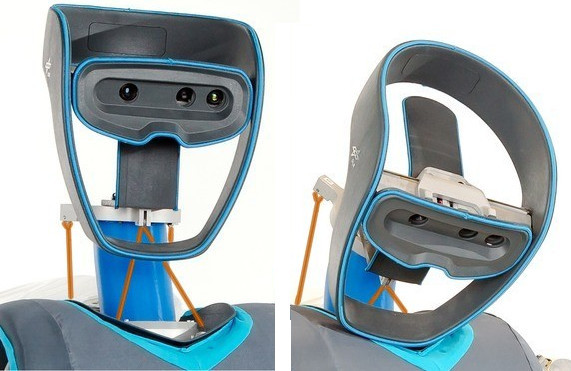}
    \subcaption{Tendon-driven elastic continuum neck in a humanoid robot}
    \label{fig:david-neck}
  \end{minipage}
  \begin{minipage}[t]{.5\linewidth}
    \centering\includegraphics[width=\linewidth]{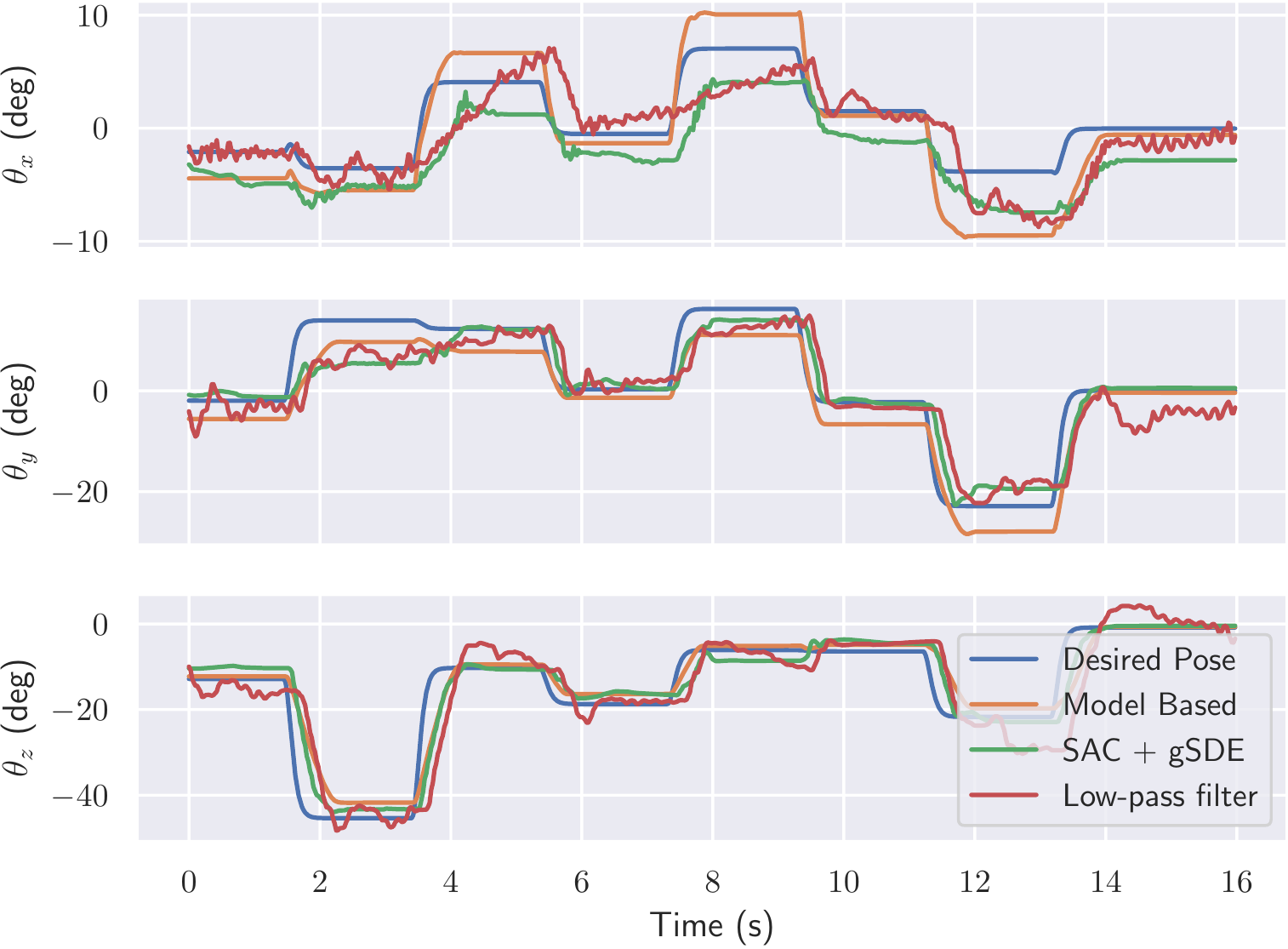}
    \subcaption{Model-Based controller vs RL controller on the real robot}
    \label{fig:eval-traj}
  \end{minipage}
  \caption{(a) The tendon-driven robot~\citep{reinecke2016structurally} used for the experiment. The tendons are highlighted in orange. (b) The model-based controller and the RL agent performs similarly on an evaluation trajectory.}
\end{figure}

\paragraph{Experiment setup}

To assess the usefulness of \ourSDE, we apply it on a real system. The task is to control a tendon-driven elastic continuum neck~\citep{reinecke2016structurally} (see~\Cref{fig:david-neck}) to a given target pose. Controlling such a soft robot is challenging because of the nonlinear tendon coupling, together with the deformation of the structure that needs to be modeled accurately. This modeling is computationally expensive~\citep{deutschmann2017position, deutschmann2019six} and requires assumptions that may not hold in the physical system.

The system is under-actuated (there are only 4 tendons), hence, the desired pose is a 4D vector: 3 angles for the rotation $\theta_x$, $\theta_y$, $\theta_z$ and one for the position $x$. The input is a 16D vector composed of: the measured tendon lengths (4D), the current tendon forces (4D), the current pose (4D) and the target pose (4D). The reward is a weighted sum between the negative geodesic distance to the desired orientation and the negative Euclidean distance to the desired position. The weights are chosen such that the two components have the same magnitude. We also add a small continuity cost to reduce the oscillations in the final policy.
The action space consists in desired delta in tendon forces, limited to \qty{5}{\newton}. For safety reasons, the tendon forces are clipped below \qty{10}{\newton} and above \qty{40}{\newton}.
An episode terminates either when the agent reaches the desired pose or after a timeout of \qty{5}{\second}.
The episode is considered successful if the desired pose is reached within a threshold of \qty{10}{\mm} for the position and \ang{5} for the orientation.
The agent controls the tendons forces at \qty{30}{\hertz}, while a PD controller monitors the motor current at \qty{3}{\kilo\hertz} on the robot. Gradient updates are directly done on a 4-core laptop, after each episode.

\paragraph{Results}

We first ran the unstructured exploration on the robot, but had to stop the experiment early: the high-frequency noise in the command was damaging the tendons and would have broken them due to their friction on the bearings.
Therefore, as a baseline, we trained a policy using \sac with a hand-crafted action smoothing (\qty{2}{\hertz} cutoff Butterworth low-pass filter) for two hours.\\
Then, we trained a controller using \sac with \ourSDE for the same duration.
We compare both learned controllers to an existing model-based controller (passivity-based approach) presented in~\citep{deutschmann2017position, deutschmann2019six} using a pre-defined trajectory (\cf~\Cref{fig:eval-traj}).
On the evaluation trajectory, the controllers are equally precise (\cf~\Cref{tab:res-eval-traj}): the mean error in orientation is below \ang{3} and the one in position below \qty{3}{\mm}.
However, the policy trained with the low-pass filter is much more jittery than the two others. We quantify this jitter as the mean absolute difference between two timesteps, denoted as \textit{continuity cost} in~\Cref{tab:res-eval-traj}.

\begin{table}[h!]
\renewcommand{\arraystretch}{1.2}
\centering
\scalebox{0.75}{
\begin{tabular}{@{}l l l l l@{}}

\toprule
& Unstructured & \ourSDE & Low-pass filter & Model-Based\\
& noise &  &  & \\
  \midrule
 Position error (mm) & N/A & 2.65 +/- 1.6 & 1.98 +/- 1.7 & 1.32 +/- 1.2 \\
 Orientation error (deg) & N/A & 2.85 +/- 2.9 & 3.53 +/- 4.0 & 2.90 +/- 2.8 \\
 Continuity cost (deg) & N/A & \textbf{0.20 +/- 0.04} & 0.38 +/- 0.07 & \textbf{0.16 +/- 0.04} \\
\bottomrule \\
\end{tabular}
}

\caption{Comparison of the mean error in position, orientation and mean continuity cost on the evaluation trajectory. We highlight best approaches when the difference is significant. The model-based and learned controllers yield comparable results but the policy trained with a low-pass filter has a much higher continuity cost.}
\label{tab:res-eval-traj}
\end{table}

\paragraph{Additional Real Robot Experiments}

\begin{figure}[h]
  \begin{minipage}[t]{.48\linewidth}
    \centering\includegraphics[width=\linewidth]{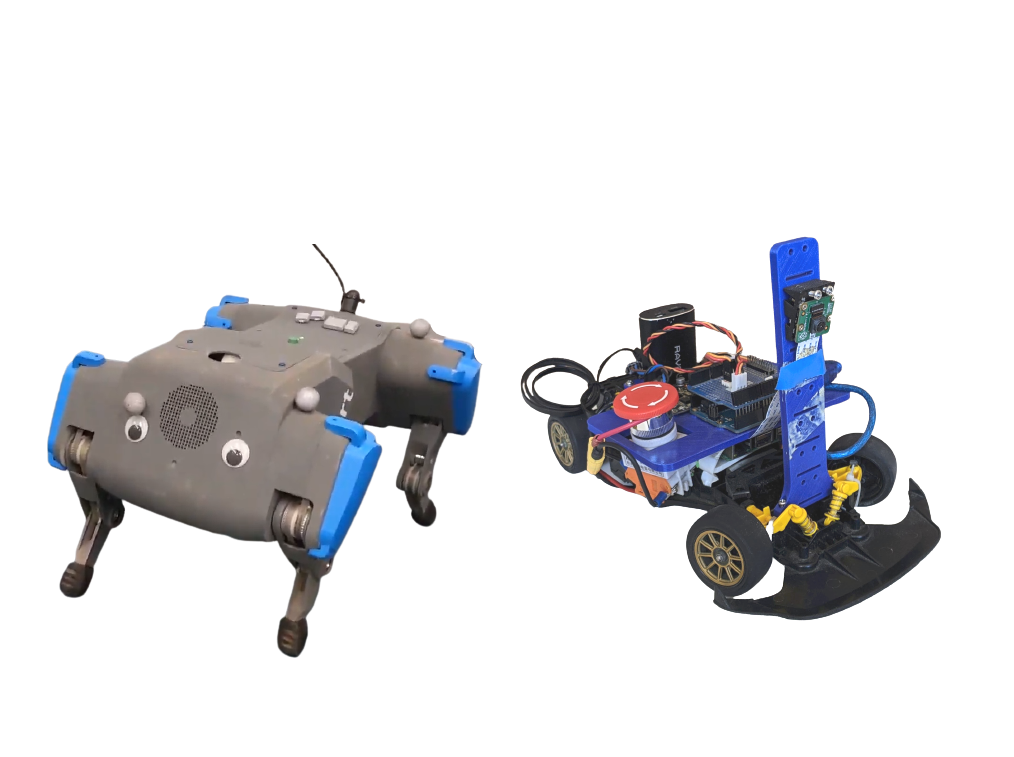}
    \subcaption{Quadruped and RC car robots}
    \label{fig:robots}
  \end{minipage}
  \begin{minipage}[t]{.48\linewidth}
    \centering\includegraphics[width=\linewidth]{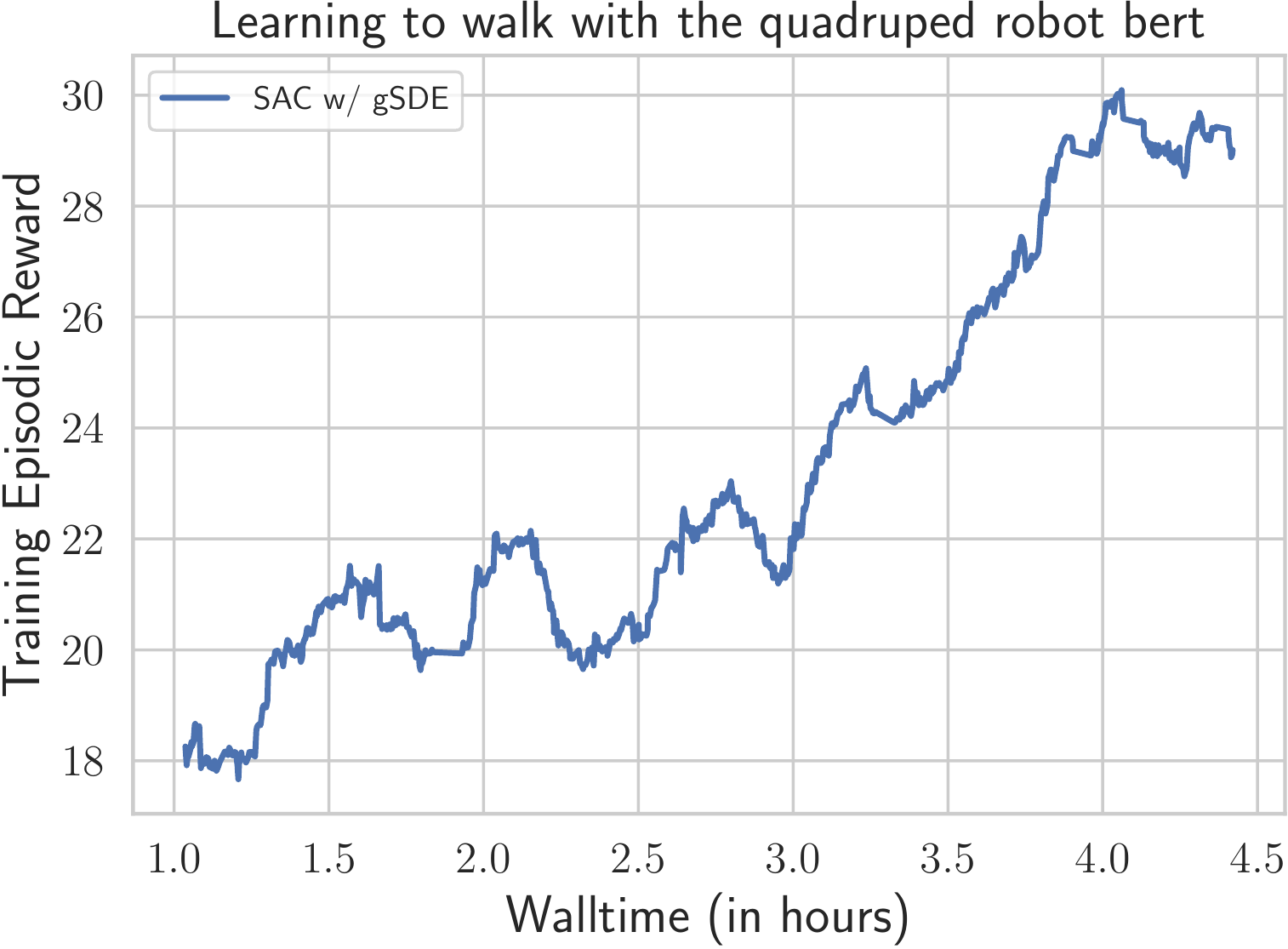}
    \subcaption{Learning curve on the walking task for the quadruped robot}
    \label{fig:training-bert}
  \end{minipage}
  \caption{Additional robots successfully trained using \sac with \ourSDE directly in the real world.}
\end{figure}

We also successfully applied \sac with \ourSDE on two additional real robotics tasks (see~\Cref{fig:robots}): training an elastic quadruped robot to walk (learning curve in~\Cref{fig:training-bert}) and learning to drive around a track using an RC car.
Both experiments are fully done on the real robot, without the use of simulation nor filter.
We provide in the supplementary material the videos of the trained controllers.

\section{Related Work}

Exploration is a key topic in reinforcement learning~\citep{sutton2018reinforcement}. It has been extensively studied in the discrete case and most recent papers still focus on discrete actions~\citep{osband2016deep, osband2018randomized}.

Several works tackle the issues of unstructured exploration for continuous control by replacing it with correlated noise. \citet{korenkevych2019autoregressive} use an autoregressive process and introduce two variables that allows to control the smoothness of the exploration. In the same vein, \citet{hoof2017generalized} rely on a temporal coherence parameter to interpolate between the step- or episode-based exploration, making use of a Markov chain to correlate the noise. This smoothed noise comes at a cost: it requires an history, which changes the problem definition.

Exploring in parameter space~\citep{kober2009policy, sehnke2010parameter, ruckstiess2010exploring, stulp2013robot, sigaud2019policy} is an orthogonal approach that also solves some issues of the unstructured exploration. It was successfully applied to real robot but relied on motor primitives~\citep{peters2008reinforcement, stulp2013robot}, which requires expert knowledge.
\citet{plappert2017parameter} adapt parameter exploration to Deep RL by defining a distance in the action space and applying layer normalization to handle high-dimensional space.

Population based algorithms, such as Evolution strategies (ES) or Genetic Algorithms (GA), also explore in parameter space. Thanks to massive parallelization, they were shown to be competitive~\citep{such2017deep} with RL in terms of training time, at the cost of being sample inefficient. To address this problem,
recent works~\citep{pourchot2018cem} proposed to combine ES exploration with RL gradient update. This combination, although powerful, unfortunately adds numerous hyperparameters and a non-negligible computational overhead.

Obtaining smooth control is essential for real robot, but it is usually overlooked by the DeepRL community.
Recently, \citet{mysore2021caps} integrated a continuity and smoothing loss inside RL algorithms.
Their approach is effective to obtain a smooth controller that reduces the energy used at test time on the real robot.
However, it does not solve the issue of smooth exploration at train time, limiting their training to simulation only.

\section{Conclusion}

In this work, we highlighted several issues that arise from the unstructured exploration in Deep RL algorithms for continuous control. Due to those issues, these algorithms cannot be directly applied to learning on real-world robots.

To address these issues, we adapt \sde to Deep RL algorithms by extending the original formulation: we sample the noise every $n$ steps and replace the exploration function input by learned features.
This generalized version (\ourSDE), provides a simple and efficient alternative to unstructured Gaussian exploration.

\ourSDE achieves very competitive results on several continuous control benchmarks, while reducing wear-and-tear at train time.
We also investigate the contribution of each modification by performing an ablation study: the noise sampling interval has the most impact and allows a compromise between performance and smoothness.
Our proposed exploration strategy, combined with \sac, is robust to hyperparameter choice, which makes it suitable for robotics applications.
To demonstrate it, we successfully apply \sac with \ourSDE directly on three different robots.

Although much progress is being made in \textit{sim2real} approaches, we believe
more effort should be invested in learning directly on real systems, even if this poses challenges in terms of safety and duration of learning.
This paper is meant as a step towards this goal, and we hope that it will revive interest in developing exploration methods that can be directly applied to real robots.

\bibliography{bibliography}  %

\clearpage

\appendix

\section{Supplementary Material}

\subsection{State Dependent Exploration}
\label{sec:sde-eq}

In the linear case, \ie with a linear policy and a noise matrix, parameter space exploration and \SDE are equivalent:

\begin{align*}
  \at &= \mu(\st; \theta_\mu) + \noise(\st; \theta_\noise), & \theta_\noise \sim \normal(0, \sigma^2) \\
      &= \theta_\mu \st + \theta_\noise \st \\
      &= (\theta_\mu + \theta_\noise) \st
\end{align*}

Because we know the policy distribution, we can obtain the derivative of the log-likelihood $\log \policy(\action|\state)$ with respect to the variance $\sigma$:

\begin{align}
  \label{eq:sde-logprob}
  \frac{\partial \log \policy(\action|\state)}{\partial \sigma_{ij}}
  &= \sum_k{\frac{\partial \log \policy_k(\action_k|\state)}{\partial \hat{\sigma_{j}}}
  \frac{\partial \hat{\sigma}_{j}}{\partial \sigma_{ij}}} \\
  &= \frac{\partial \log \policy_j(\action_j|\state)}{\partial \hat{\sigma}_{j}}
  \frac{\partial \hat{\sigma}_{j}}{\partial \sigma_{ij}} \\
  &= \frac{(\action_j - \mu_j)^2 - \hat{\sigma_j}^2}{\hat{\sigma_j}^3}
  \frac{\state_i^2 \sigma_{ij}}{\hat{\sigma_{j}}}
\end{align}

This can be easily plugged into the likelihood ratio gradient estimator~\citep{williams1992simple}, which allows adapting $\sigma$ during training.
\SDE is therefore compatible with standard policy gradient methods, while addressing most shortcomings of the unstructured exploration.

\subsection{Algorithms}
\label{sec:algorithms}

In this section, we shortly present the algorithms used in this paper. They correspond to state-of-the-art methods in model-free RL for continuous control, either in terms of sample efficiency or wall-clock time.

\paragraph{\aac}
\aac is the synchronous version of Asynchronous Advantage Actor-Critic (A3C)~\citep{mnih2016asynchronous}.
It is an actor-critic method that uses parallel rollouts of $n$-steps to update the policy. It relies on the \textsc{REINFORCE}~\citep{williams1992simple} estimator to compute the gradient. \aac is fast but not sample efficient.

\paragraph{\ppo}
\aac gradient update does not prevent large changes that lead to huge drop in performance.
To tackle this issue, Trust Region Policy Optimization (TRPO)~\citep{schulman2015trust} introduces a trust-region in the policy parameter space, formulated as a constrained optimization problem: it updates the policy while being close in terms of KL divergence to the old policy.
Its successor, Proximal Policy Optimization (\ppo)~\citep{schulman2017proximal} relaxes the constraints (which requires costly conjugate gradient step) by clipping the objective using importance ratio. \ppo makes also use of workers (as in \aac) and Generalized Advantage Estimation (GAE)~\citep{schulman2015high} for computing the advantage.

\paragraph{\tddd}
Deep Deterministic Policy Gradient (DDPG)~\citep{lillicrap2015continuous} combines the deterministic policy gradient algorithm~\citep{silver2014deterministic} with the improvements from Deep Q-Network (DQN)~\citep{mnih2013playing}: using a replay buffer and target networks to stabilize training.
Its direct successor, Twin Delayed DDPG (\tddd)~\citep{fujimoto2018addressing} brings three major tricks to tackle issues coming from function approximation: clipped double Q-Learning (to reduce overestimation of the Q-value function), delayed policy update (so the value function converges first) and target policy smoothing (to prevent overfitting).
Because the policy is deterministic, DDPG and \tddd rely on external noise for exploration.

\paragraph{\sac}
Soft Actor-Critic~\citep{haarnoja2017reinforcement}, successor of Soft Q-Learning (SQL)~\citep{haarnoja2018soft} optimizes the maximum-entropy objective, that is slightly different compared to the classic RL objective:
\begin{align}
  \label{eq:maxent_objective}
  J(\policy)  = \sum_{t=0}^{T} \E{(\st, \at) \sim \rho_\policy}{\reward(\st,\at) + \alpha\ent(\policy(\voidarg|\st))}.
\end{align}
where $\ent$ is the policy entropy and $\alpha$ is the entropy temperature and allows to have a trade-off between the two objectives.

\sac learns a stochastic policy, using a squashed Gaussian distribution, and incorporates the clipped double Q-learning trick from \tddd.
In its latest iteration~\citep{haarnoja2018applications}, \sac automatically adjusts the entropy coefficient $\alpha$, removing the need to tune this crucial hyperparameter.

\paragraph{Which algorithm for robotics?}

\aac and \ppo are both on-policy algorithms and can be easily parallelized, resulting in relatively small training time.
On the other hand, \sac and \tddd are off-policy and run on a single worker, but are much more sample efficient than the two previous methods, achieving equivalent performances with a fraction of the samples.

Because we are focusing on robotics applications, having multiple robots is usually not possible, which makes \tddd and \sac the methods of choice. Although \tddd and \sac are very similar, \sac embeds the exploration directly in its objective function, making it easier to tune. We also found, during our experiments in simulation, that \sac works for a wide range of hyperparameters. As a result, we adopt that algorithm for the experiment on a real robot and for the ablation study.

\subsection{Real Robot Experiments}

\paragraph{Common Setup}
For each real robot experiment, to improve smoothness of the final controller and tackle communication delays (which would break Markov assumption), we augment the input with the previous observation and the last action taken and add a small continuity cost to the reward.
For each task, we decompose the reward function into a primary (what we want to achieve) and secondary component (soft constraints such as continuity cost).
Each reward term is normalized, which allows to easily weight each component depending on their importance.
Compared to previous work, we use the exact same algorithm as the one used for simulated tasks and therefore avoid the use of filter.

\paragraph{Learning to control an elastic neck.}
An episode terminates either when the agent reaches the desired pose or after a timeout of 5s, \ie each episode has a maximum length of 200 steps. The episode is considered successful if the desired pose is reached within a threshold of $10 mm$ for the position and $5 deg$ for the orientation.

\paragraph{Learning to walk with the elastic quadruped robot bert.}
The agent receives joint angles, velocities, torques and IMU data as input (over Wi-Fi) and commands the desired absolute motor angles.
The primary reward is the distance traveled and the secondary reward is a weighted sum of different costs: heading cost, distance to the center line and continuity cost.
Thanks to a treadmill, the reset of the robot was semi-automated.
Early stopping and monitoring of the robot was done using external tracking, but the observation is computed from on-board sensors only.
An episode terminates if the robot falls, goes out of bounds or after a timeout of 5s.
Training is done directly with the real robot over several days, totalizing around 8 hours of interaction.

\paragraph{Learning to drive with a RC car.}
The agent receives an image from the on-board camera as input and commands desired throttle and steering angle.
Features are computed using a pre-trained auto-encoder as in~\citet{drive-smoothly-in-minutes}.
The primary reward is a weighted sum between a survival bonus (no intervention by the safety driver) and the commanded throttle.
There is only the continuity cost as secondary reward.
One episode terminates when the safety driver intervenes (crash) or after a timeout of 1 minute.
Training is done directly on the robot and requires less than 30 minutes of interaction.

\begin{figure}[h!]
  \centering\includegraphics[width=0.8\linewidth]{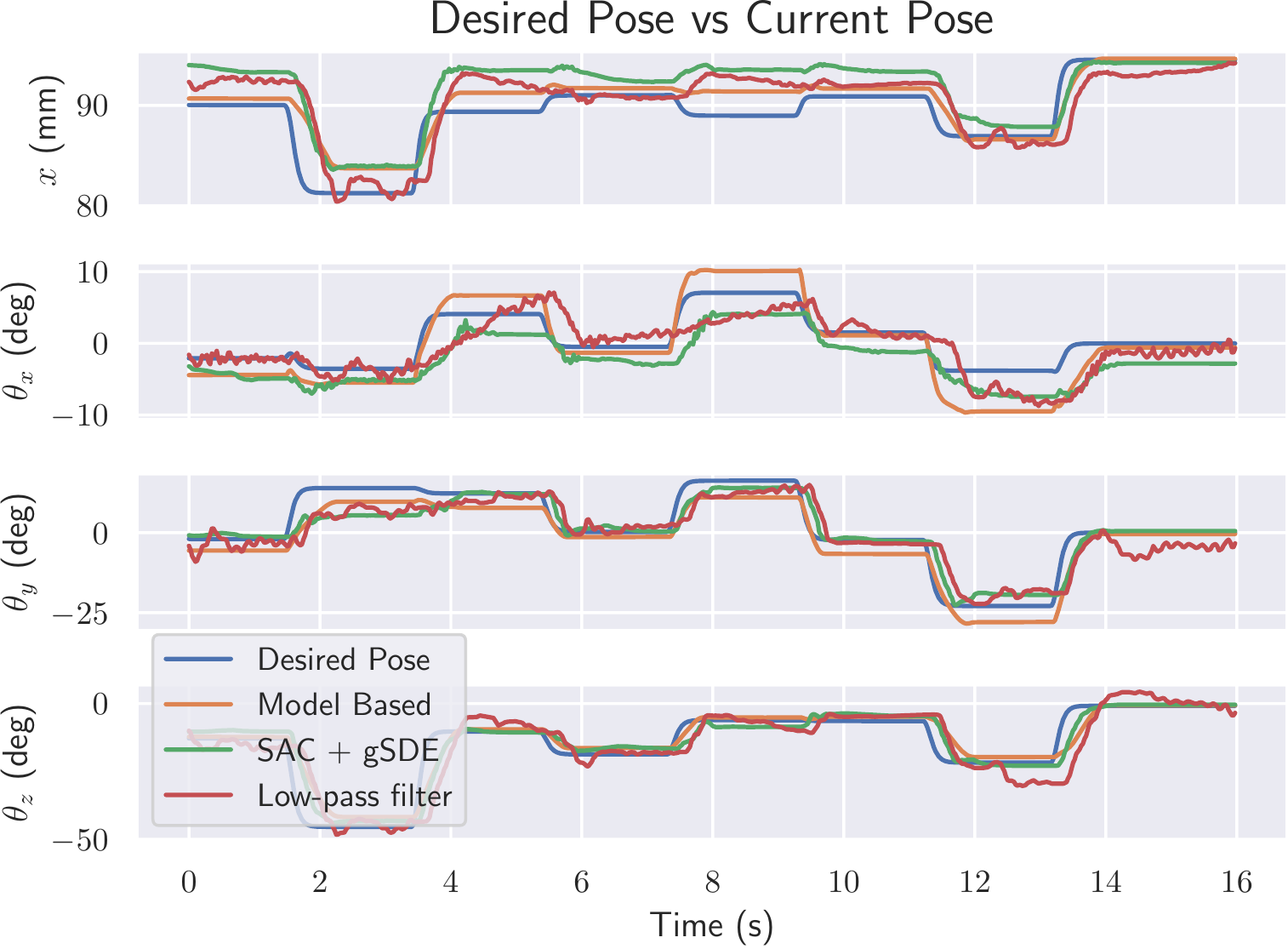}
  \label{fig:eval-traj-big}
  \caption{Comparison of the model-based controller with the learned RL agent on an evaluation trajectory: the two performs similarly.}
\end{figure}

\begin{table}[h!]
\renewcommand{\arraystretch}{1.2}
\centering
\begin{tabular}{@{}l l c l@{}}

\toprule
& \sac + \ourSDE & \phantom{abc} & Model-Based~\citep{deutschmann2017position}\\
  \midrule
 Error in position (mm) & 2.65 +/- 1.6 & & 1.32 +/- 1.2 \\
 Error in orientation (deg) & 2.85 +/- 2.9 & & 2.90 +/- 2.8 \\
\bottomrule \\
\end{tabular}

\caption{Comparison of the mean error in position and orientation on the evaluation trajectory. The model-based and learned controller yield comparable results.}
\label{tab:res-eval-traj}
\end{table}

\subsection{Implementation Details}
\label{sec:implementation}

We used a PyTorch~\citep{raffin2019baselines3} version of Stable-Baselines~\citep{hill2018stable} library, together with the RL Zoo training framework~\citep{raffin2020zoo3}.
It uses the common implementations tricks for \ppo~\citep{engstrom2020implementation} for the version using independent Gaussian noise.

For \sac, to ensure numerical stability, we clip the mean to be in range $[-2, 2]$, as it was causing infinite values. In the original implementation, a regularization $\mathcal{L}_2$ loss on the mean and standard deviation was used instead. The algorithm for \sac with \ourSDE is described in~\Cref{algo:sac-sde}.

Compared to the original \SDE paper, we did not have to use the \textit{expln} trick~\citep{ruckstiess2008state} to avoid exploding variance for PyBullet tasks. However, we found it useful on specific environment like \textit{BipedalWalkerHardcore-v2}. The original \sac implementation clips this variance.

\begin{algorithm}[h]
\caption{Soft Actor-Critic with \ourSDE}
\label{algo:sac-sde}
\begin{algorithmic}
\State \mbox{Initialize parameters} $\theta_\mu$, $\theta_Q$, $\sigma$, $\alpha$
\State \mbox{Initialize replay buffer} $\mathcal{D}$
\For{each iteration}
  \State $\theta_\noise \sim \normal(0, \sigma^2)$ \Comment{Sample noise function parameters}
	\For{each environment step}
  	\State $\at = \policy(\st) = \mu(\st; \theta_\mu) + \noise(\st; \theta_\noise)$  \Comment{Get the noisy action}
  	\State $\state_{t+1} \sim p(\state_{t+1}| \st, \at)$ \Comment{Step in the environment}
  	\State $\mathcal{D} \leftarrow \mathcal{D} \cup \left\{(\st, \at, \reward(\st, \at), \state_{t+1})\right\}$
    \Comment{Update the replay buffer}
	\EndFor
	\For{each gradient step}
    \State $\theta_\noise \sim \normal(0, \sigma^2)$ \Comment{Sample noise function parameters}
    \State Sample a minibatch from the replay buffer $\mathcal{D}$
    \State Update the entropy temperature $\alpha$
    \State Update parameters using $\nabla J_Q$ and $\nabla J_\policy$
    \Comment{Update actor $\mu$, critic $Q$ and noise variance $\sigma$}
    \State Update target networks
	\EndFor
\EndFor
\end{algorithmic}
\end{algorithm}

\subsection{Learning Curves and Additional Results}
\label{sec:learning-curves}

\begin{figure}[h!]
  \centering\includegraphics[width=0.5\linewidth]{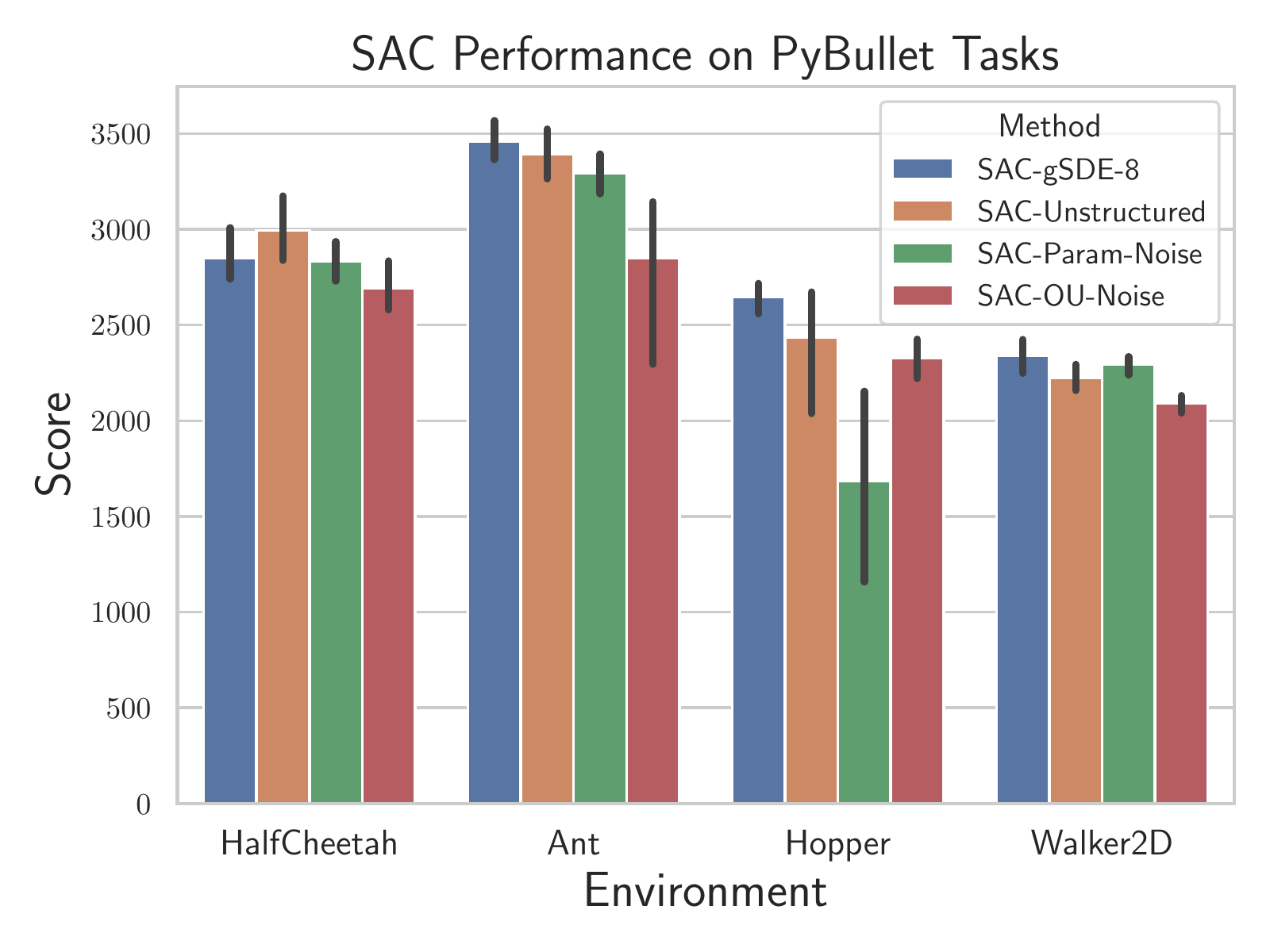}
  \label{fig:sac-pybullet-return}
  \caption{Performance results for SAC with different type of exploration on PyBullet tasks}
\end{figure}

\begin{table}[h!]
\vspace{1em}
\renewcommand{\arraystretch}{1}
\centering
\scalebox{.8}{
\begin{tabular}{@{}l ll c ll@{}}

\toprule
 & \multicolumn{2}{c}{\textbf{\aac}} & \phantom{abc} & \multicolumn{2}{c}{\textbf{\ppo}} \\
 \cmidrule{2-3} \cmidrule{5-6}

 Environments & \multicolumn{1}{c}{\ourSDE} & \multicolumn{1}{c}{Gaussian} && \multicolumn{1}{c}{\ourSDE} &  \multicolumn{1}{c}{Gaussian} \\ \midrule
 \hc & \textbf{2028} +/- \textbf{107} & 1652 +/- 94 && \textbf{2760} +/- \textbf{52} & 2254 +/- 66 \\
 \ant & \textbf{2560} +/- \textbf{45} & 1967 +/- 104 && \textbf{2587} +/- \textbf{133} & 2160 +/- 63 \\
 \hopper & 1448 +/- 163 & 1559 +/- 129 && \textbf{2508} +/- \textbf{16} & 1622 +/- 220 \\
 \walker & \textbf{694} +/- \textbf{73} & 443 +/- 59 && \textbf{1776} +/- \textbf{53} & 1238 +/- 75 \\

\bottomrule \\
\end{tabular}
}

\caption{Final performance (higher is better) of \aac and \ppo on 4 environments with \ourSDE and unstructured Gaussian exploration. We report the mean over 10 runs of 2 million steps.
For each benchmark, we highlight the results of the method with the best mean, when the difference is statistically significant.
}
\label{tab:res-bullet-onpolicy}
\end{table}

\begin{table}[h!]
\renewcommand{\arraystretch}{1}
\centering
\scalebox{.8}{
\begin{tabular}{@{}l ll c ll@{}}

\toprule
 & \multicolumn{2}{c}{\textbf{\sac}} & \phantom{abc} & \multicolumn{2}{c}{\textbf{\tddd}} \\
 \cmidrule{2-3} \cmidrule{5-6}

 Environments & \multicolumn{1}{c}{\ourSDE} & \multicolumn{1}{c}{Gaussian} && \multicolumn{1}{c}{\ourSDE}  &  \multicolumn{1}{c}{Gaussian} \\ \midrule
 \hc & 2850 +/- 73 & 2994 +/- 89 && 2578 +/- 44 & 2687 +/- 67 \\
 \ant & 3459 +/- 52 & 3394 +/- 64 && \textbf{3267} +/- \textbf{34} & 2865 +/- 278 \\
 \hopper & 2646 +/- 45 & 2434 +/- 190 && 2353 +/- 78 & 2470 +/- 111 \\
 \walker & \textbf{2341} +/- \textbf{45} & 2225 +/- 35 && 1989 +/- 153 & 2106 +/- 67 \\

\bottomrule \\
\end{tabular}
}
\caption{Final performance (higher is better) of \sac and \tddd on 4 environments with \ourSDE and unstructured Gaussian exploration. We report the mean over 10 runs of 1 million steps. For each benchmark, we highlight the results of the method with the best mean, when the difference is statistically significant.
}
\label{tab:res-bullet-offpolicy}
\end{table}

\begin{figure}[h]
  \begin{minipage}[t]{.5\linewidth}
    \centering\includegraphics[width=\linewidth]{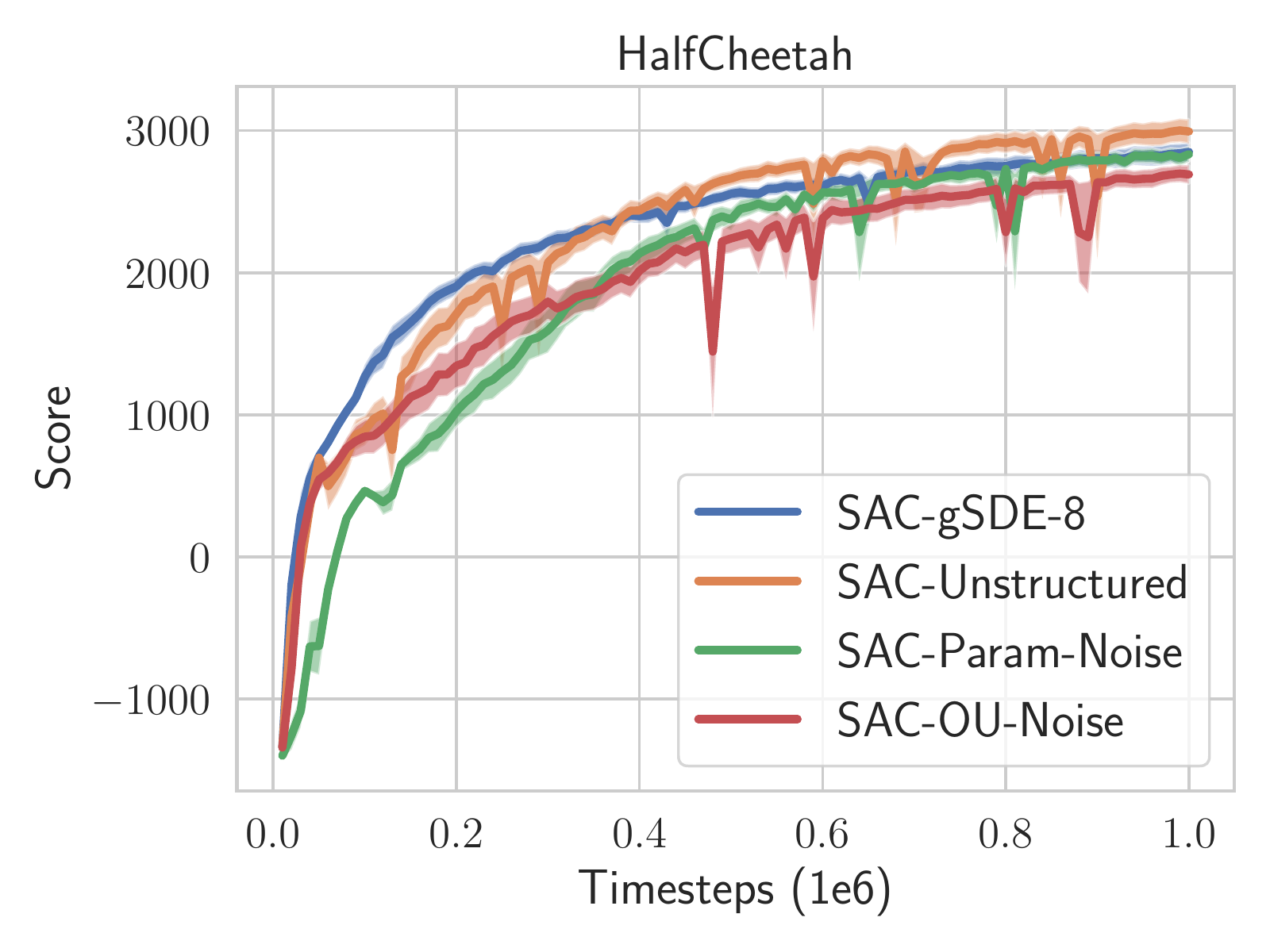}
    \subcaption{\hc}
  \end{minipage}
  \begin{minipage}[t]{.5\linewidth}
    \centering\includegraphics[width=\linewidth]{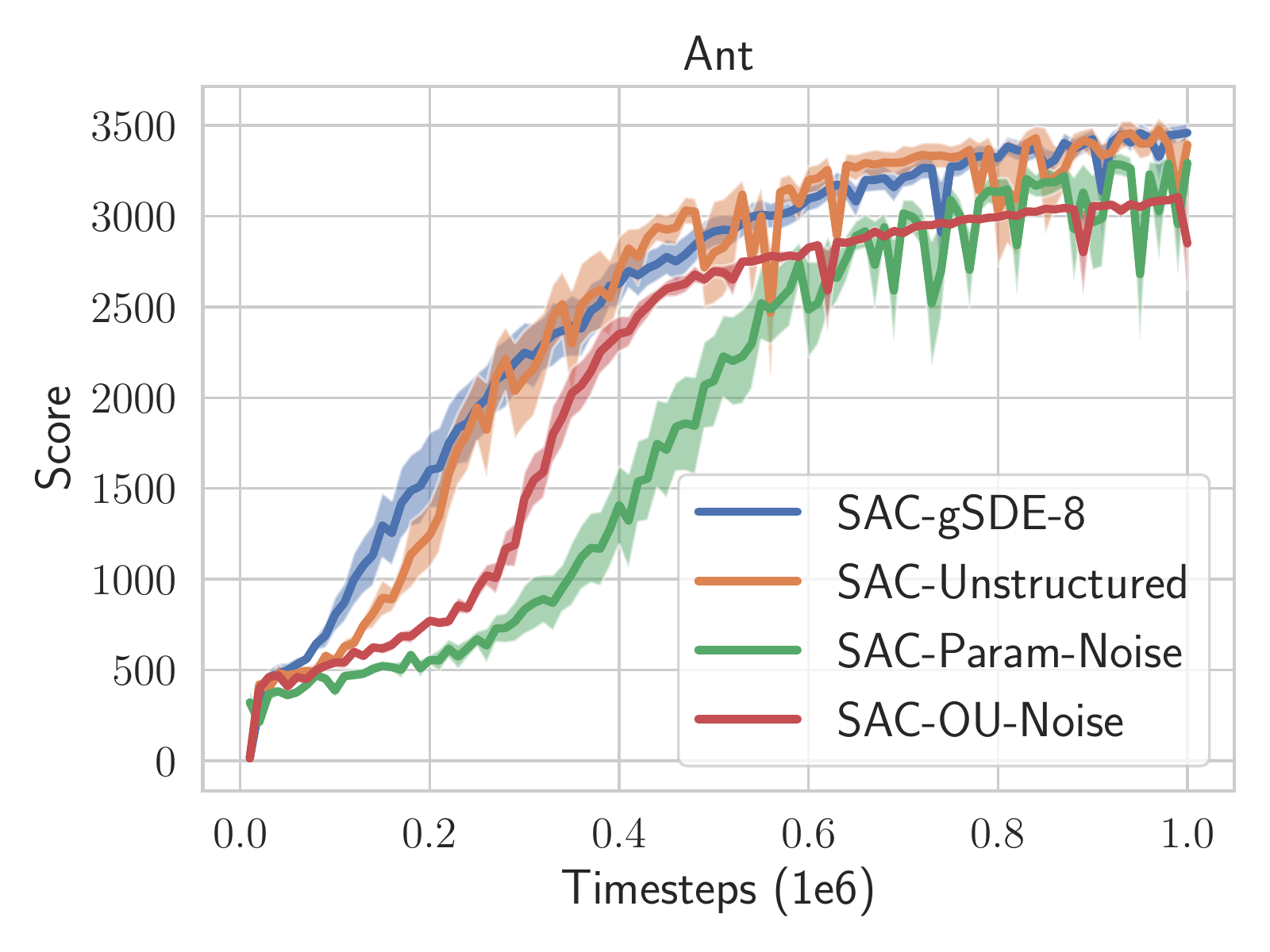}
    \subcaption{\ant}
  \end{minipage}
  \begin{minipage}[t]{.5\linewidth}
    \centering\includegraphics[width=\linewidth]{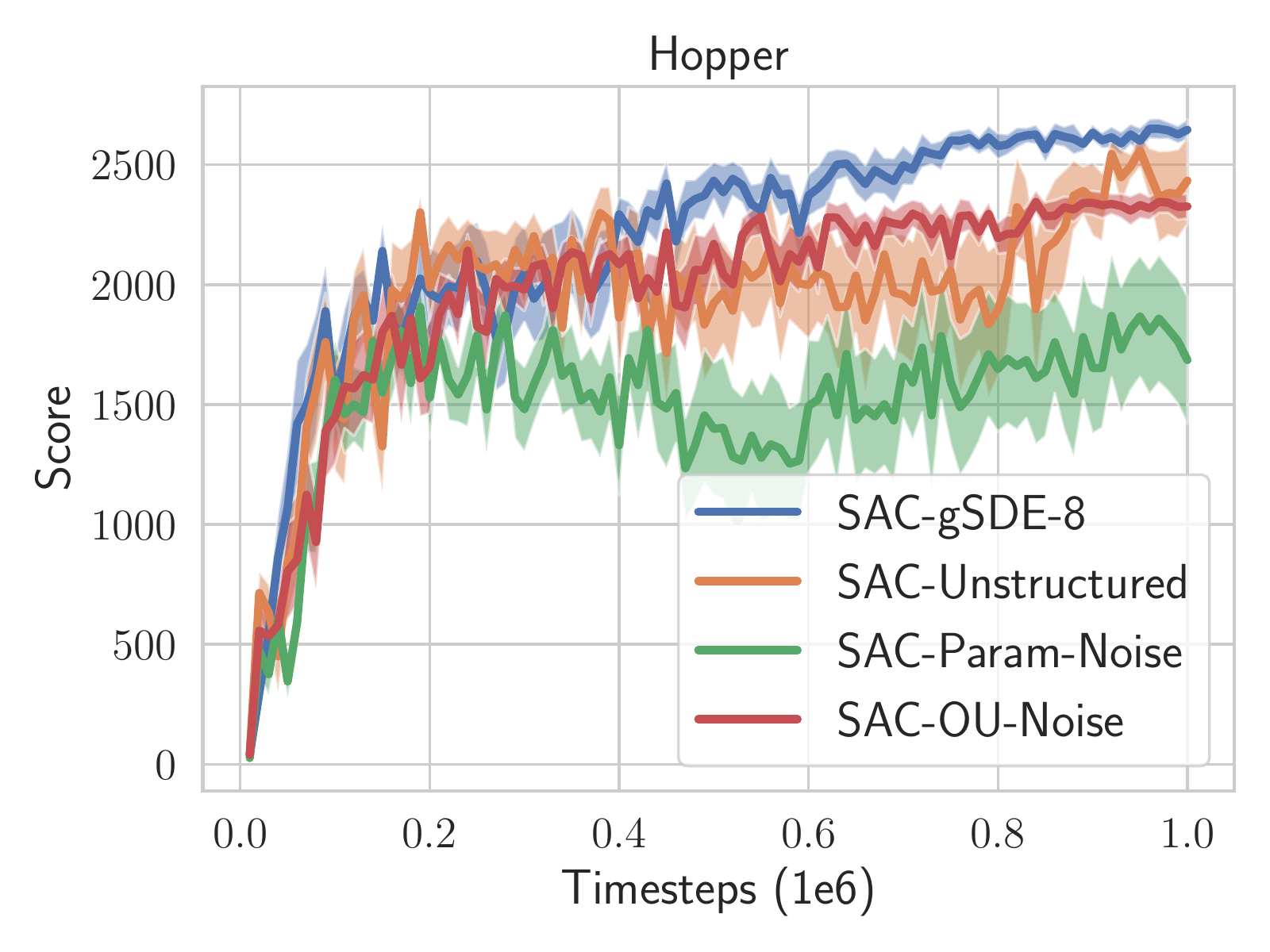}
    \subcaption{\hopper}
  \end{minipage}
  \begin{minipage}[t]{.5\linewidth}
    \centering\includegraphics[width=\linewidth]{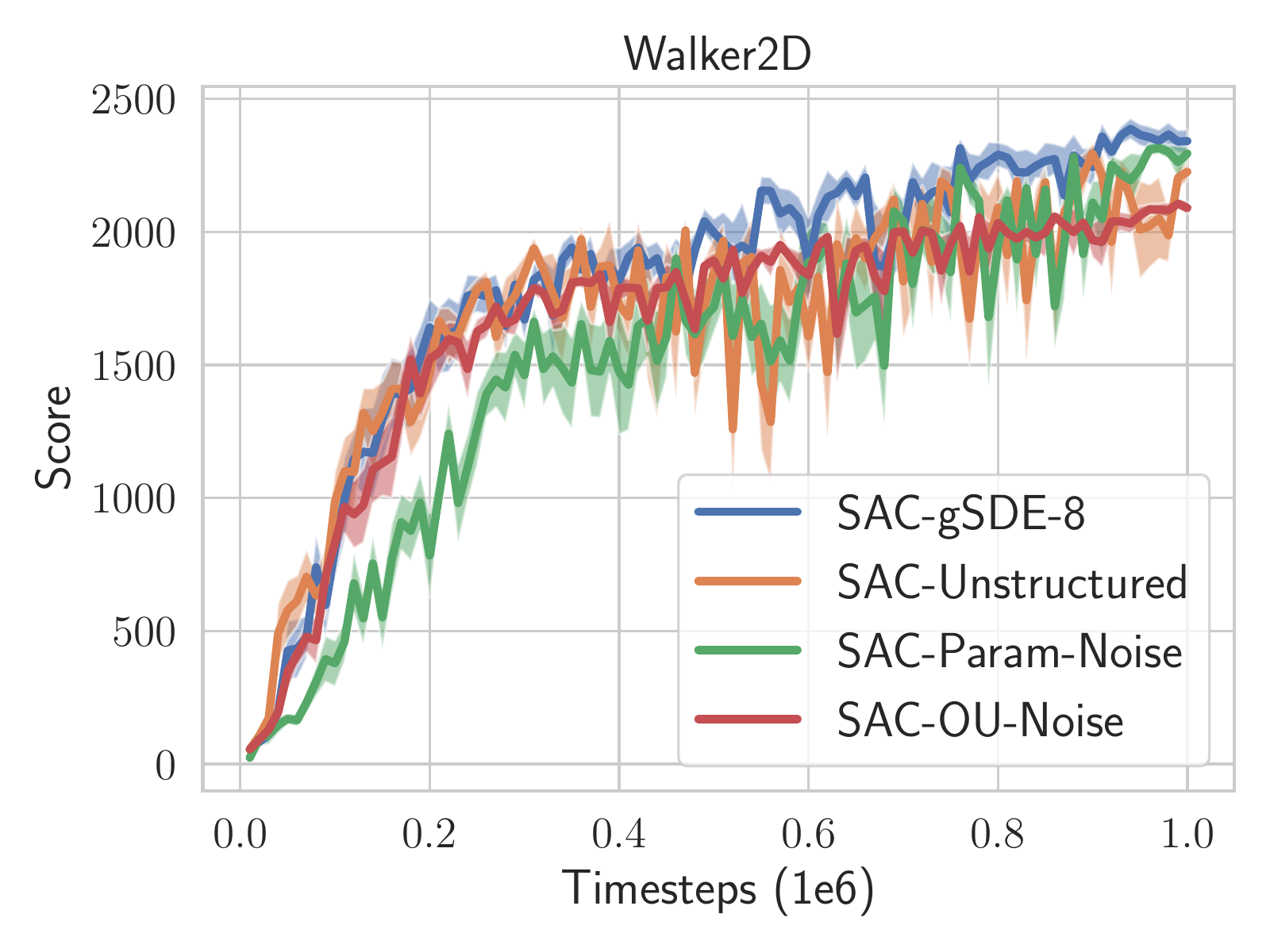}
    \subcaption{\walker}
  \end{minipage}
  \caption{Learning curves for SAC with different type of exploration on PyBullet tasks. The line denotes the mean over 10 runs of 1 million steps.}
  \label{fig:learning-curves-sac}
\end{figure}

\Cref{fig:learning-curves-sac} shows the learning curves for \sac with different types of exploration noise.

\begin{figure}[h]
  \begin{minipage}[t]{.5\linewidth}
    \centering\includegraphics[width=\linewidth]{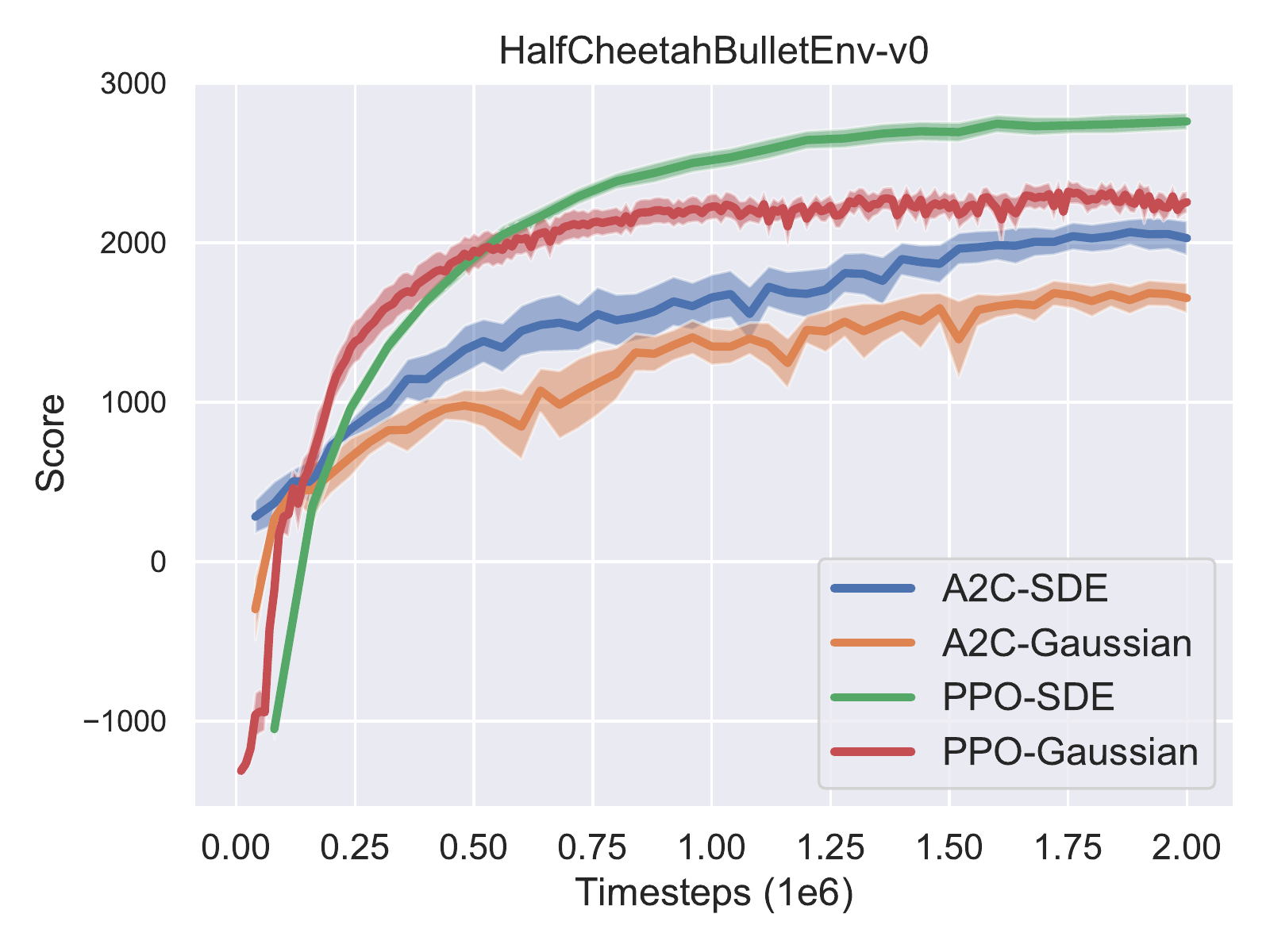}
    \subcaption{\hc}
  \end{minipage}
  \begin{minipage}[t]{.5\linewidth}
    \centering\includegraphics[width=\linewidth]{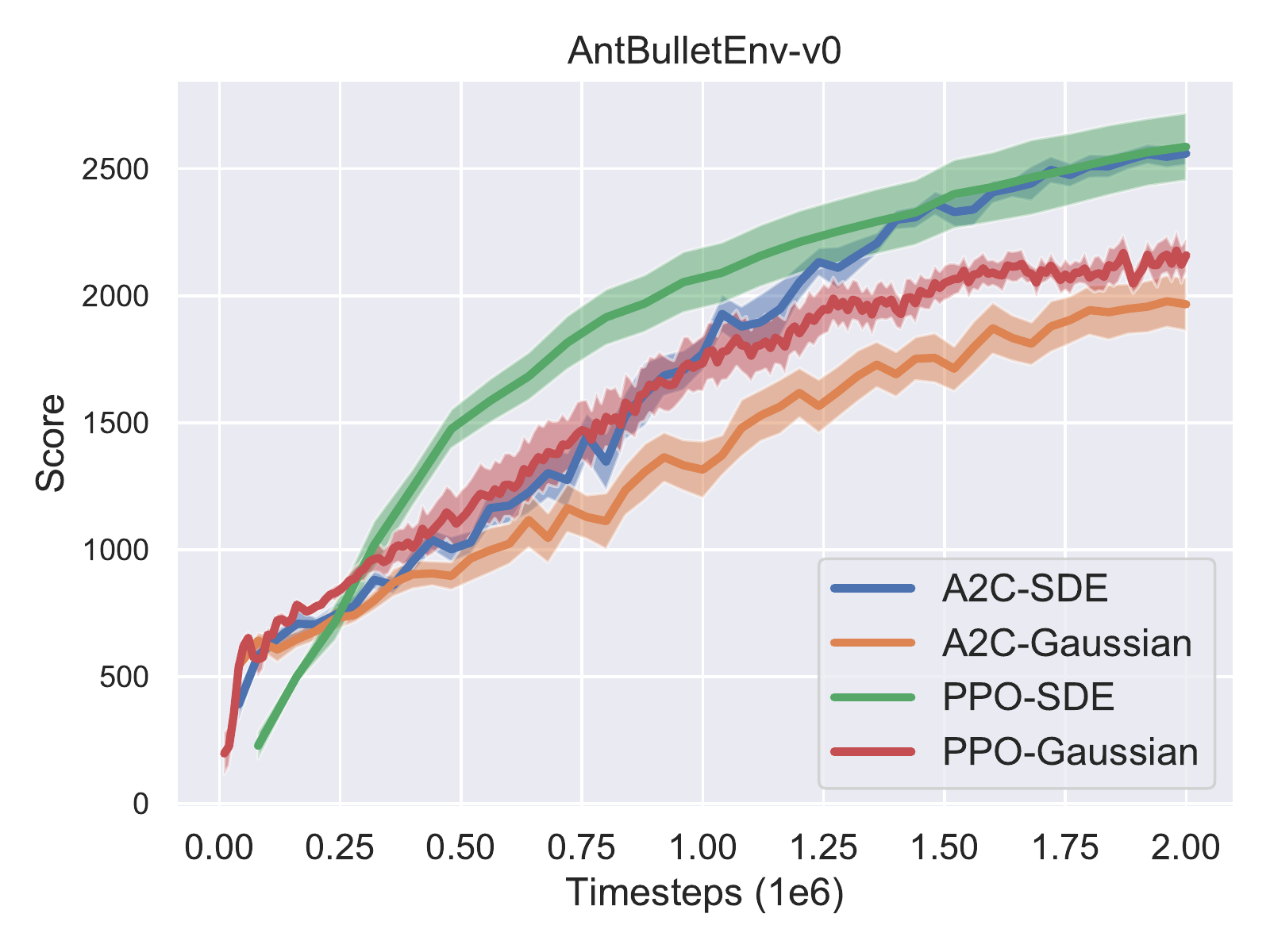}
    \subcaption{\ant}
  \end{minipage}
  \begin{minipage}[t]{.5\linewidth}
    \centering\includegraphics[width=\linewidth]{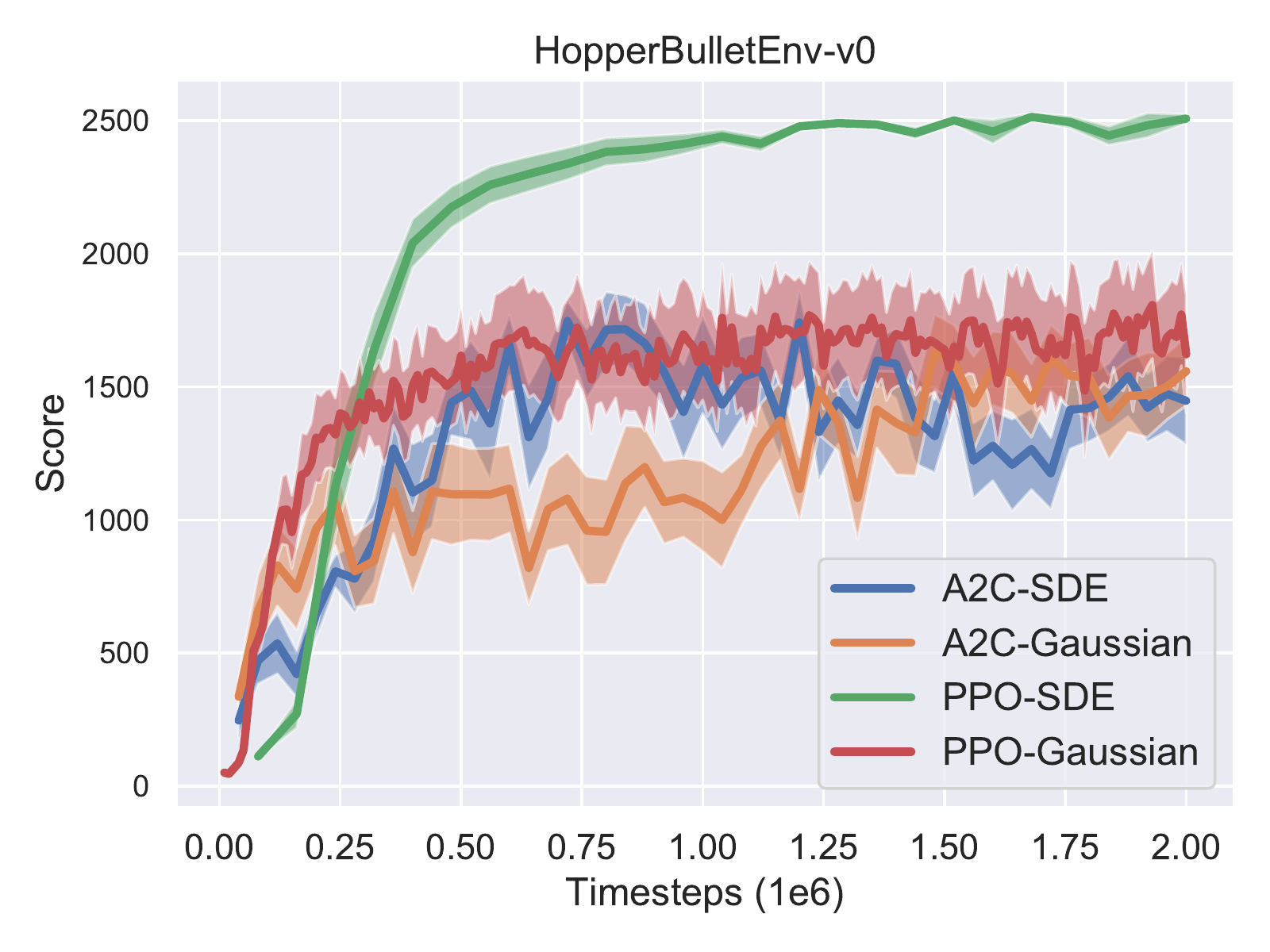}
    \subcaption{\hopper}
  \end{minipage}
  \begin{minipage}[t]{.5\linewidth}
    \centering\includegraphics[width=\linewidth]{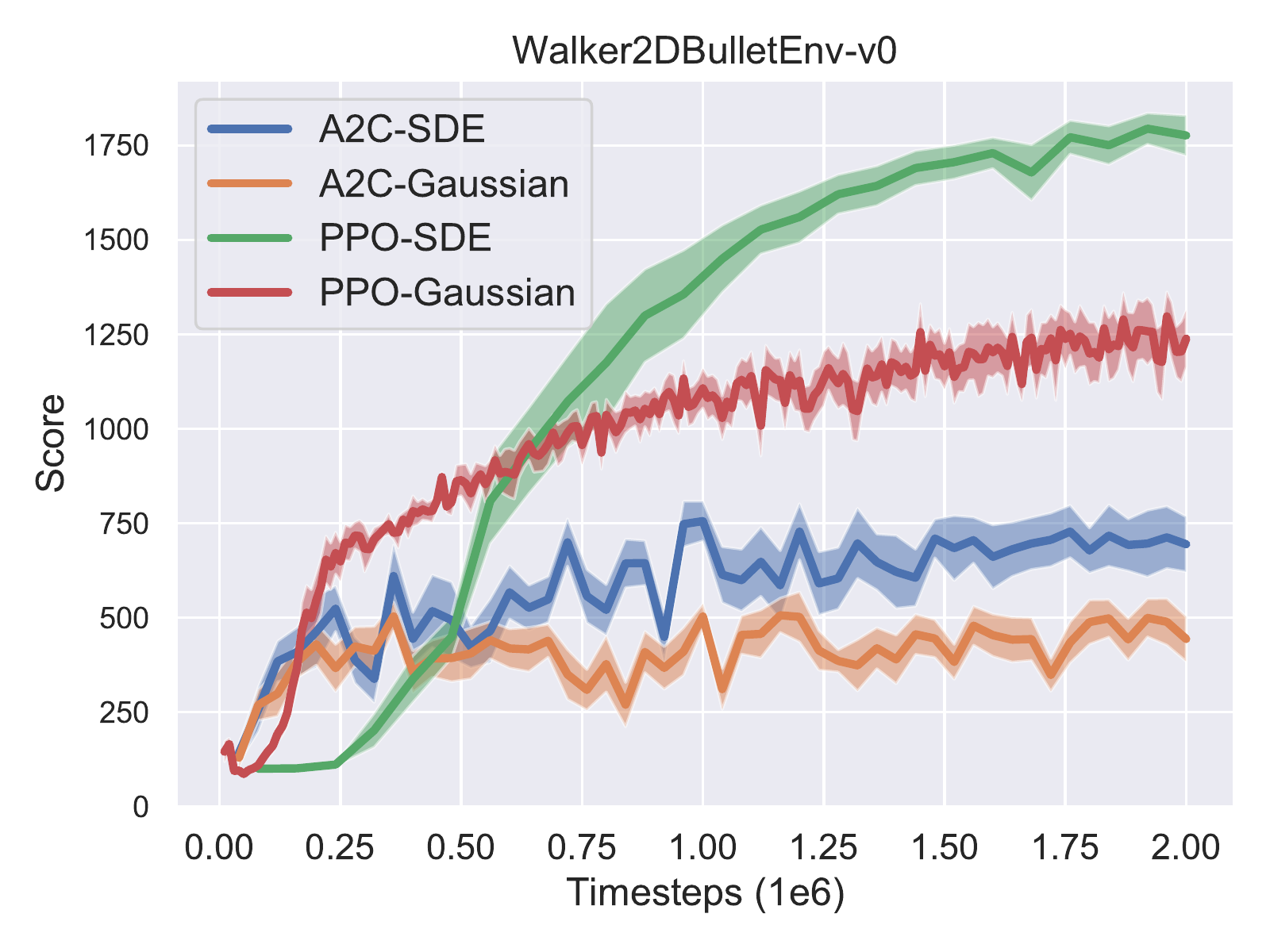}
    \subcaption{\walker}
  \end{minipage}
  \caption{Learning curves for on-policy algorithms on PyBullet tasks. The line denotes the mean over 10 runs of 2 million steps.}
  \label{fig:learning-curves-onpolicy}
\end{figure}

\begin{figure}[h]
  \begin{minipage}[t]{.5\linewidth}
    \centering\includegraphics[width=\linewidth]{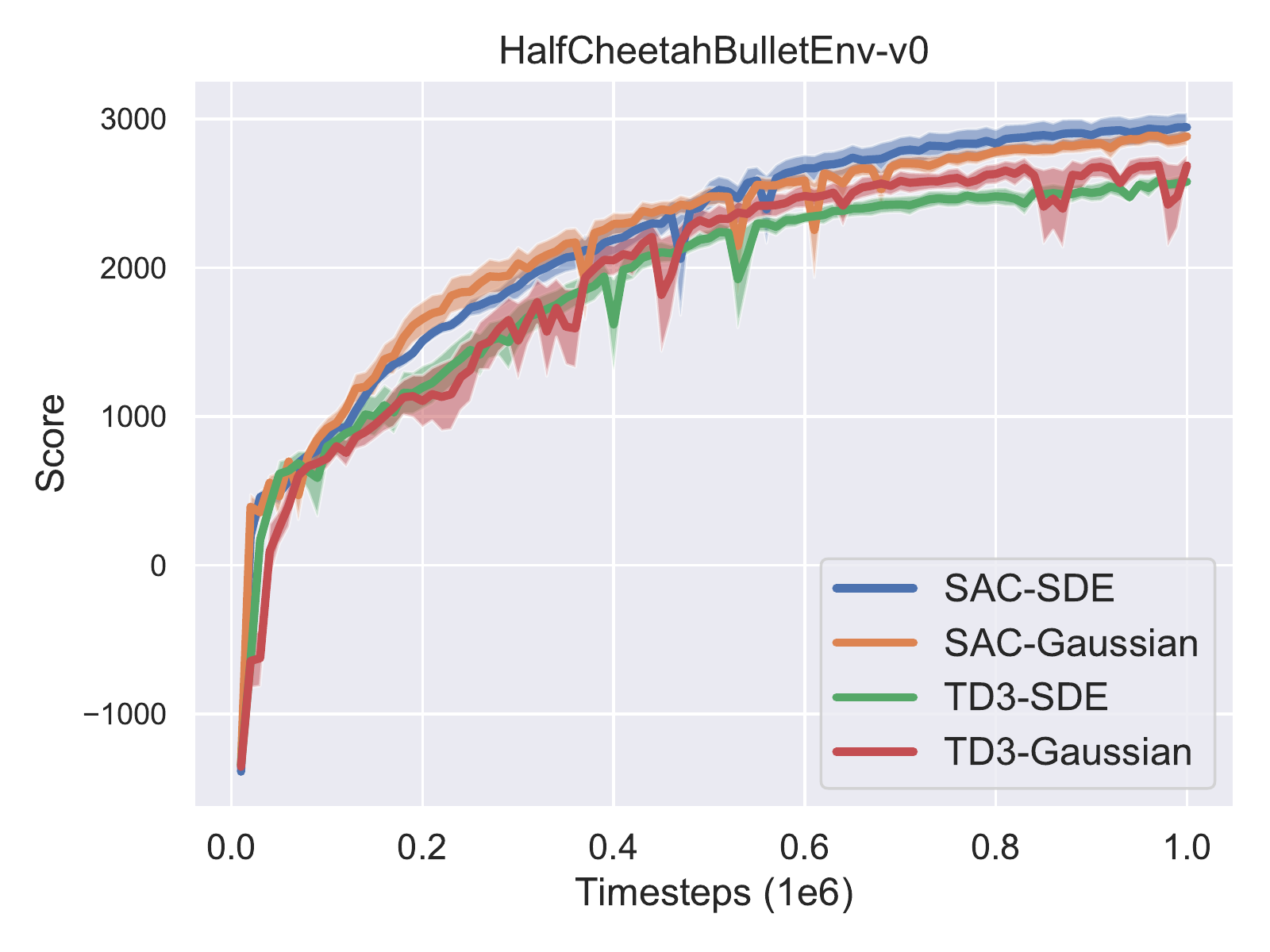}
    \subcaption{\hc}
  \end{minipage}
  \begin{minipage}[t]{.5\linewidth}
    \centering\includegraphics[width=\linewidth]{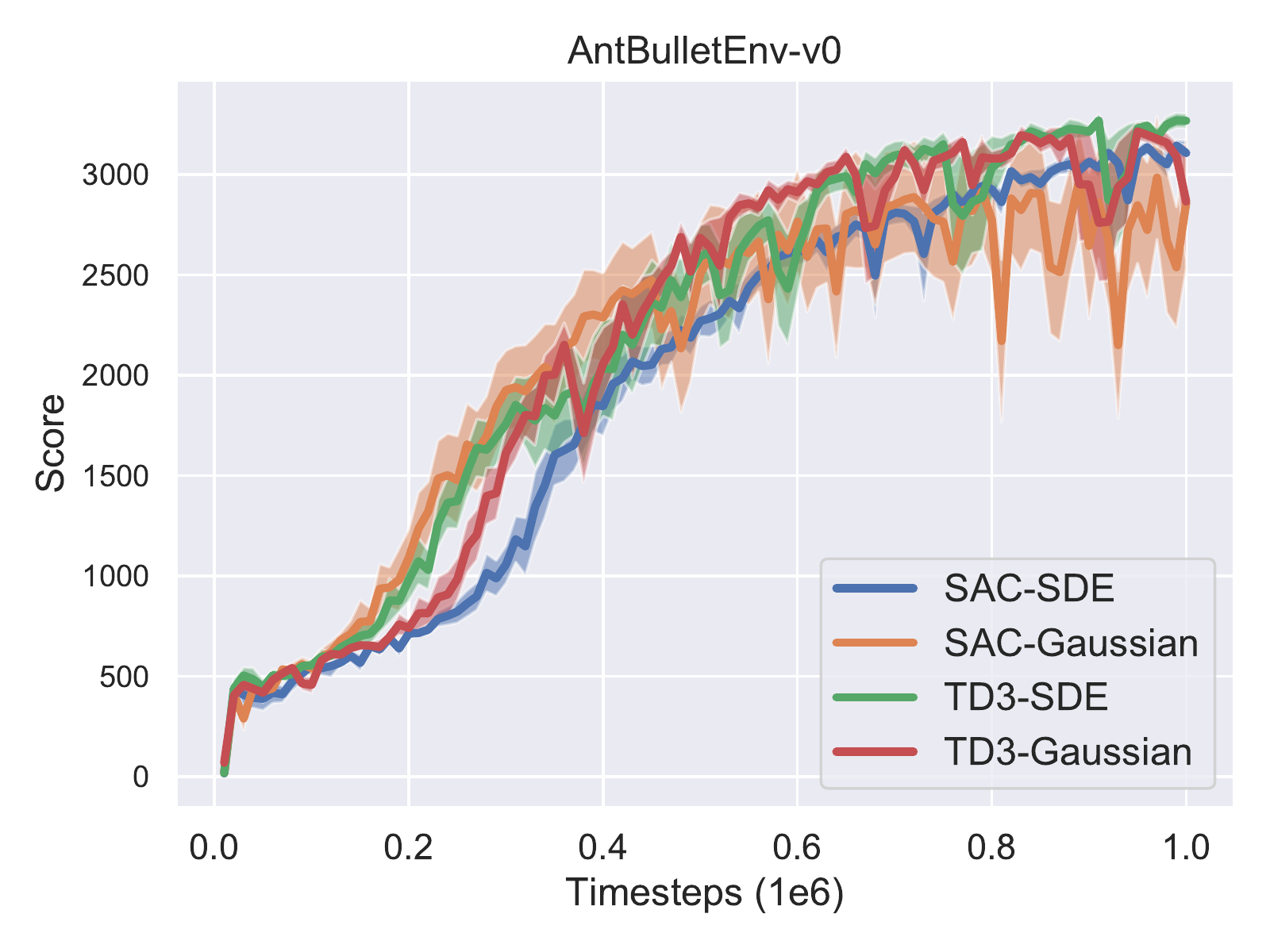}
    \subcaption{\ant}
  \end{minipage}
  \begin{minipage}[t]{.5\linewidth}
    \centering\includegraphics[width=\linewidth]{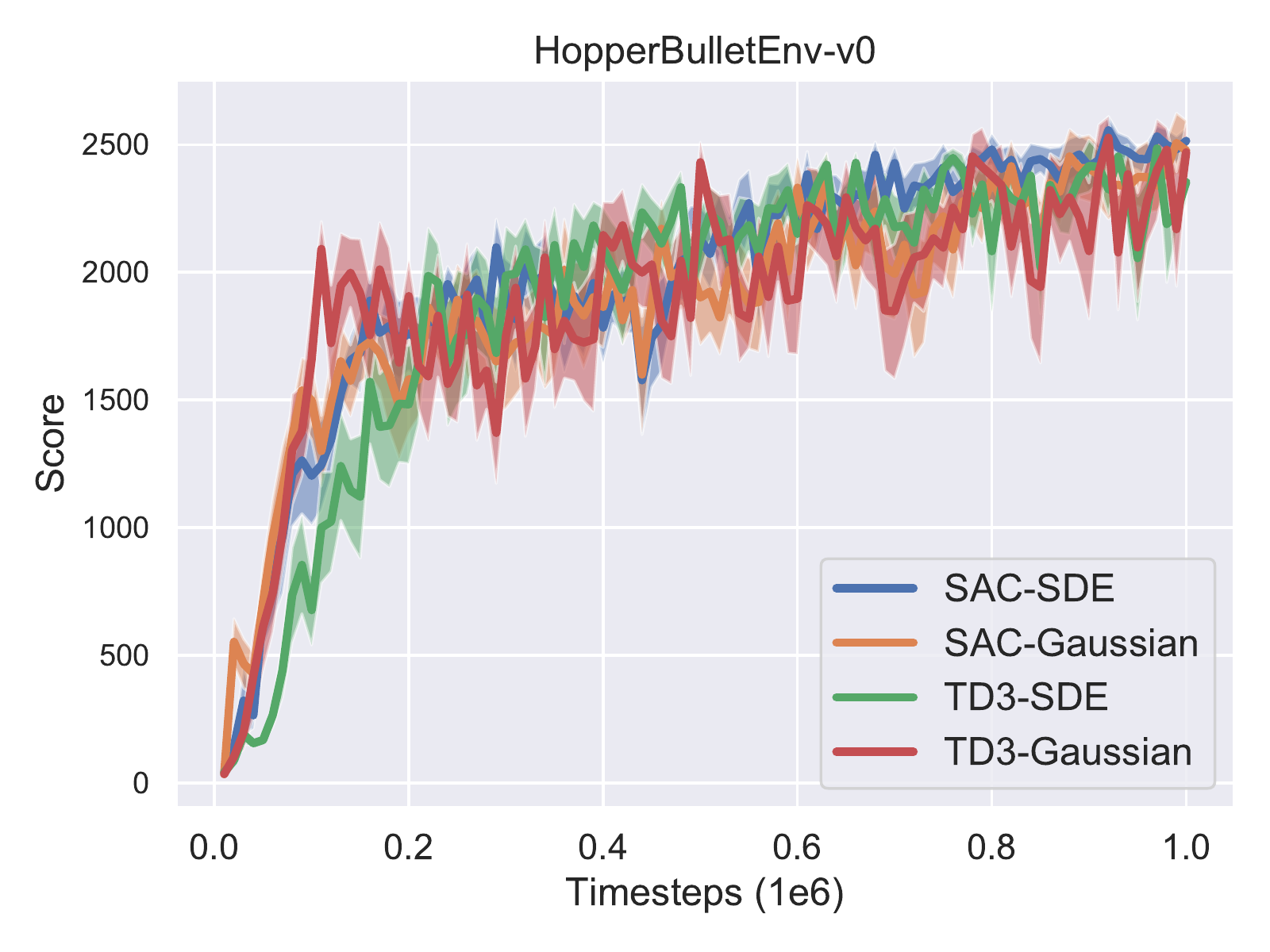}
    \subcaption{\hopper}
  \end{minipage}
  \begin{minipage}[t]{.5\linewidth}
    \centering\includegraphics[width=\linewidth]{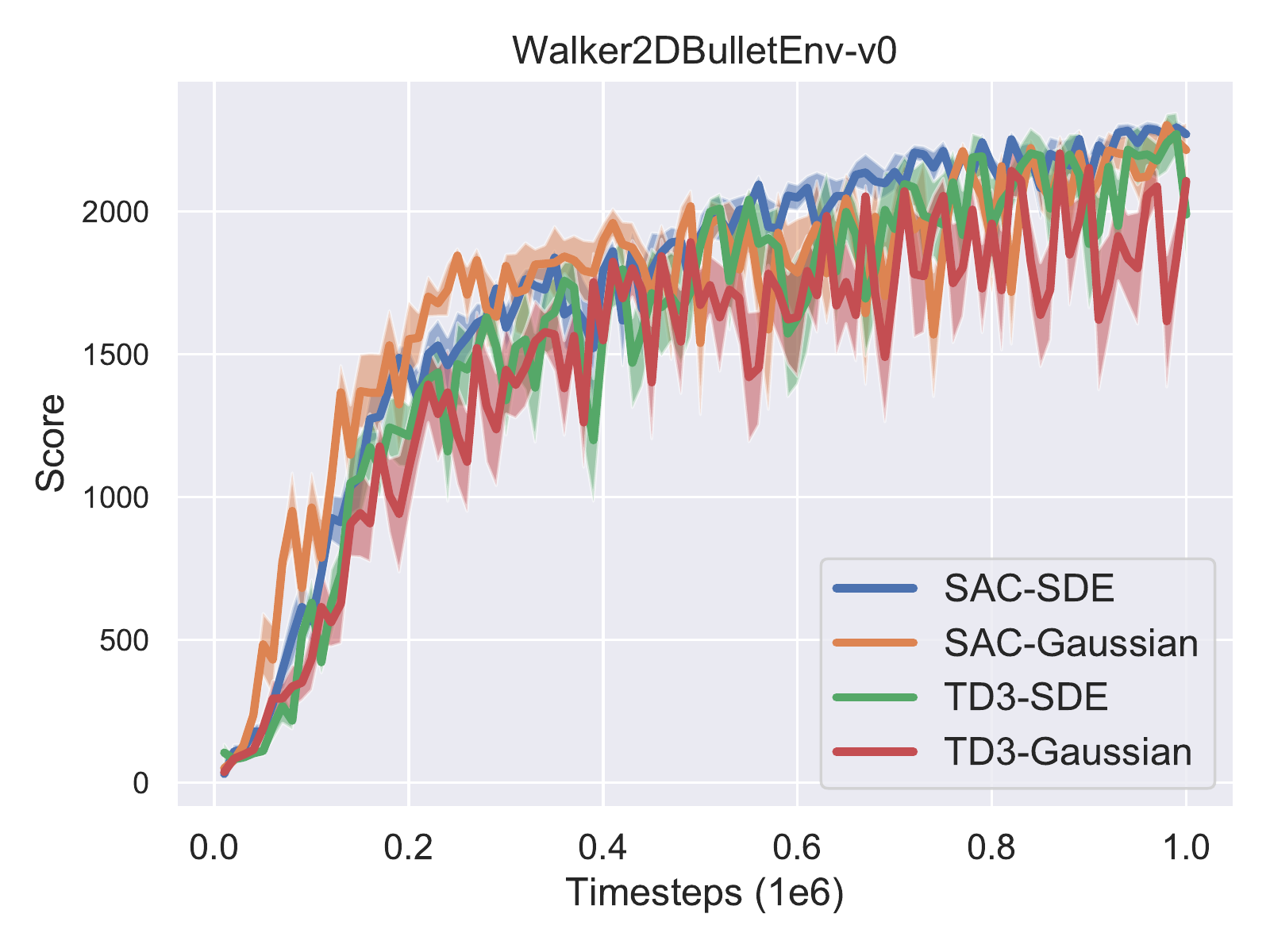}
    \subcaption{\walker}
  \end{minipage}
  \caption{Learning curves for off-policy algorithms on PyBullet tasks. The line denotes the mean over 10 runs of 1 million steps.}
  \label{fig:learning-curves-offpolicy}
\end{figure}

\Cref{fig:learning-curves-onpolicy} and \Cref{fig:learning-curves-offpolicy} show the learning curves for off-policy and on-policy algorithms on the four PyBullet tasks, using \ourSDE or unstructured Gaussian exploration.

\subsection{Ablation Study: Additional Plots}
\label{appendix:ablation}

\begin{figure}[h]
  \begin{minipage}[t]{.5\linewidth}
    \centering\includegraphics[width=\linewidth]{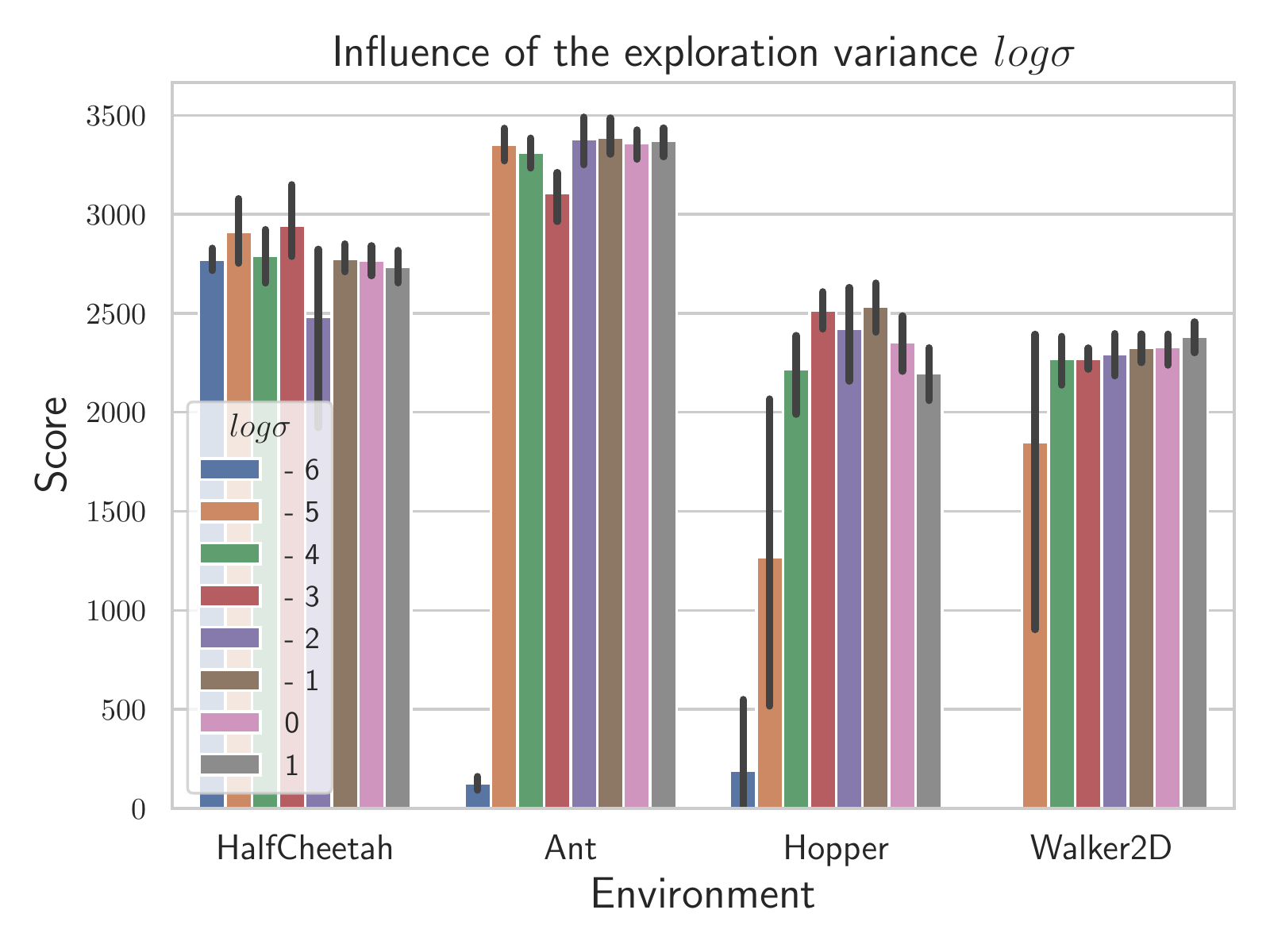}
    \subcaption{Initial exploration variance $\log \sigma$ (\sac)}
  \end{minipage}
  \begin{minipage}[t]{.5\linewidth}
    \centering\includegraphics[width=\linewidth]{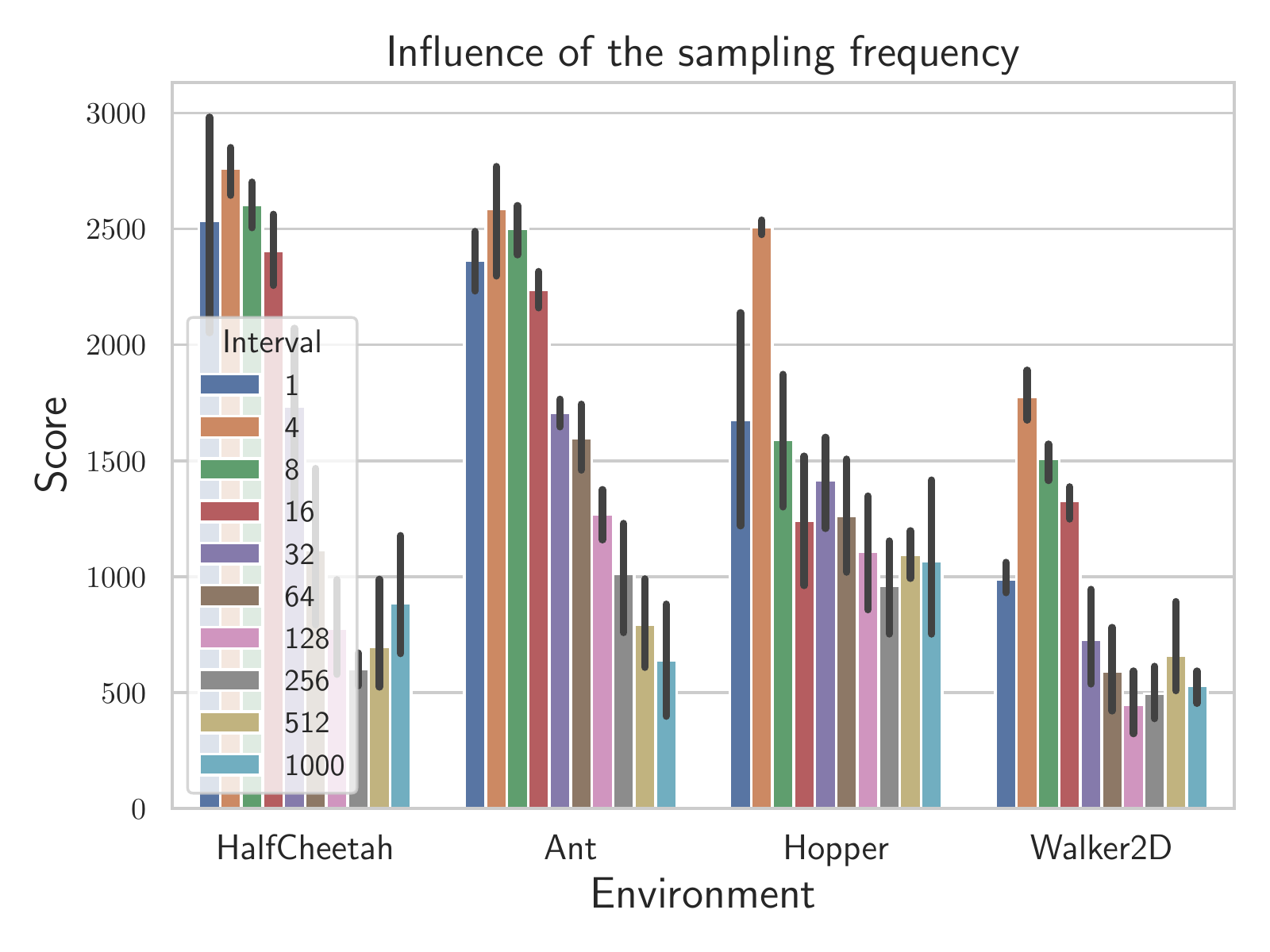}
    \subcaption{Noise sampling interval (\ppo)}
  \end{minipage}
  \caption{Sensitivity of \sac and \ppo to selected hyperparameters on PyBullet tasks}
  \label{fig:ablation-additional}
\end{figure}

\Cref{fig:ablation-additional} displays the ablation study on remaining PyBullet tasks. It shows that \sac is robust against initial exploration variance, and \ppo results highly depend on the noise sampling interval.

\begin{figure}[h]
  \begin{minipage}[t]{.45\linewidth}
    \centering\includegraphics[width=\linewidth]{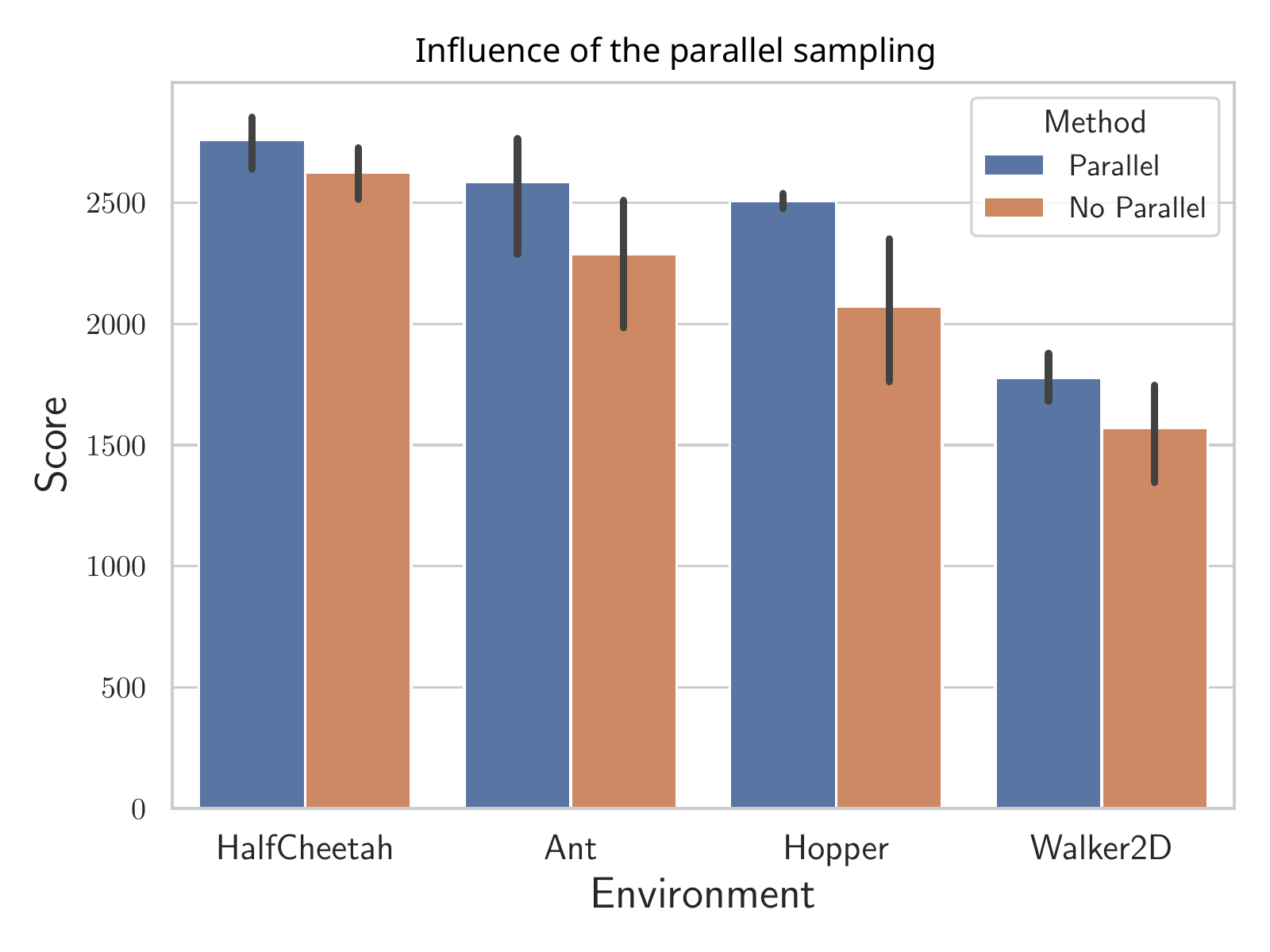}
    \subcaption{Effect of parallel sampling for \ppo}
    \label{fig:ppo-parallel}
  \end{minipage}
  \hfill
  \begin{minipage}[t]{.45\linewidth}
    \centering\includegraphics[width=\linewidth]{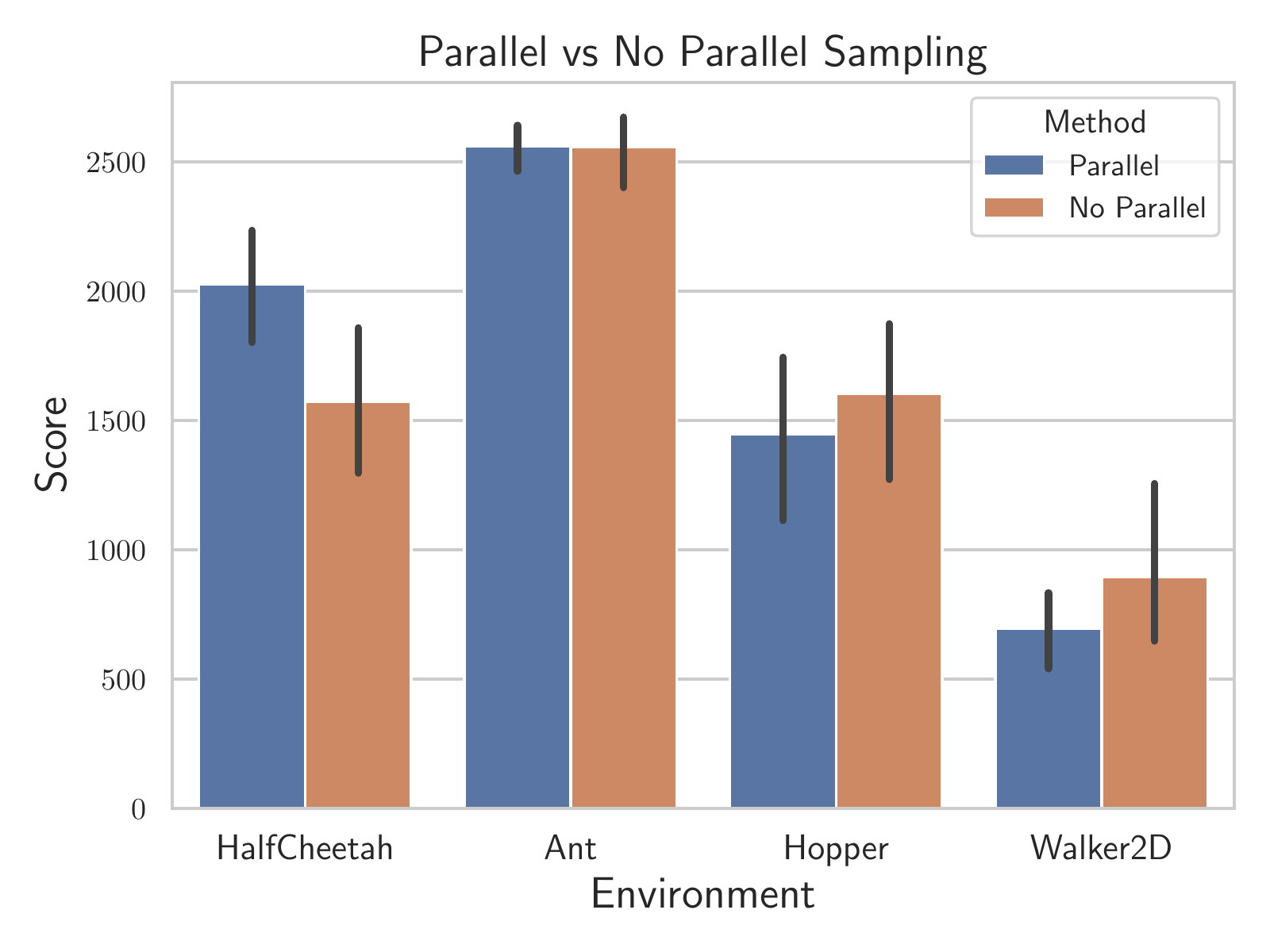}
    \subcaption{Effect of parallel sampling for \aac}
    \label{fig:a2c-parallel}
  \end{minipage}
  \caption{Parallel sampling of the noise matrix has a positive impact for \ppo (a) and \aac (b) on PyBullet tasks.}
  \label{fig:ablation-parallel}
\end{figure}

\paragraph{Parallel Sampling}
The effect of sampling a set of noise parameters per worker is shown for \ppo in~\Cref{fig:ppo-parallel}. This modification improves the performance for each task, as it allows a more diverse exploration. Although less significant, we observe the same outcome for \aac on PyBullet environments (\cf~\Cref{fig:a2c-parallel}).
Thus, making use of parallel workers improves both exploration and the final performance.

\subsection{Hyperparameter Optimization}
\label{sec:optim}

\ppo and \tddd hyperparameters for unstructured exploration are reused from the original papers~\citep{schulman2017proximal, fujimoto2018addressing}.
For \sac, the optimized hyperparameters for \ourSDE are performing better than the ones from \citet{haarnoja2018soft}, so we keep them for the other types of exploration to have a fair comparison.
No hyperparameters are available for \aac in \citet{mnih2016asynchronous} so we use the tuned one from \citet{raffin2018zoo}.

To tune the hyperparameters, we use a TPE sampler and a median pruner from Optuna~\citep{akiba2019optuna} library.
We give a budget of 500 candidates with a maximum of $3 \cdot 10^{5}$ time-steps on the \hc environment. Some hyperparameters are then manually adjusted (\eg increasing the replay buffer size) to improve the stability of the algorithms.

\subsection{Hyperparameters}
\label{sec:hyperparameters}

\begin{figure}[h]
  \begin{minipage}[t]{.5\linewidth}
    \centering\includegraphics[width=\linewidth]{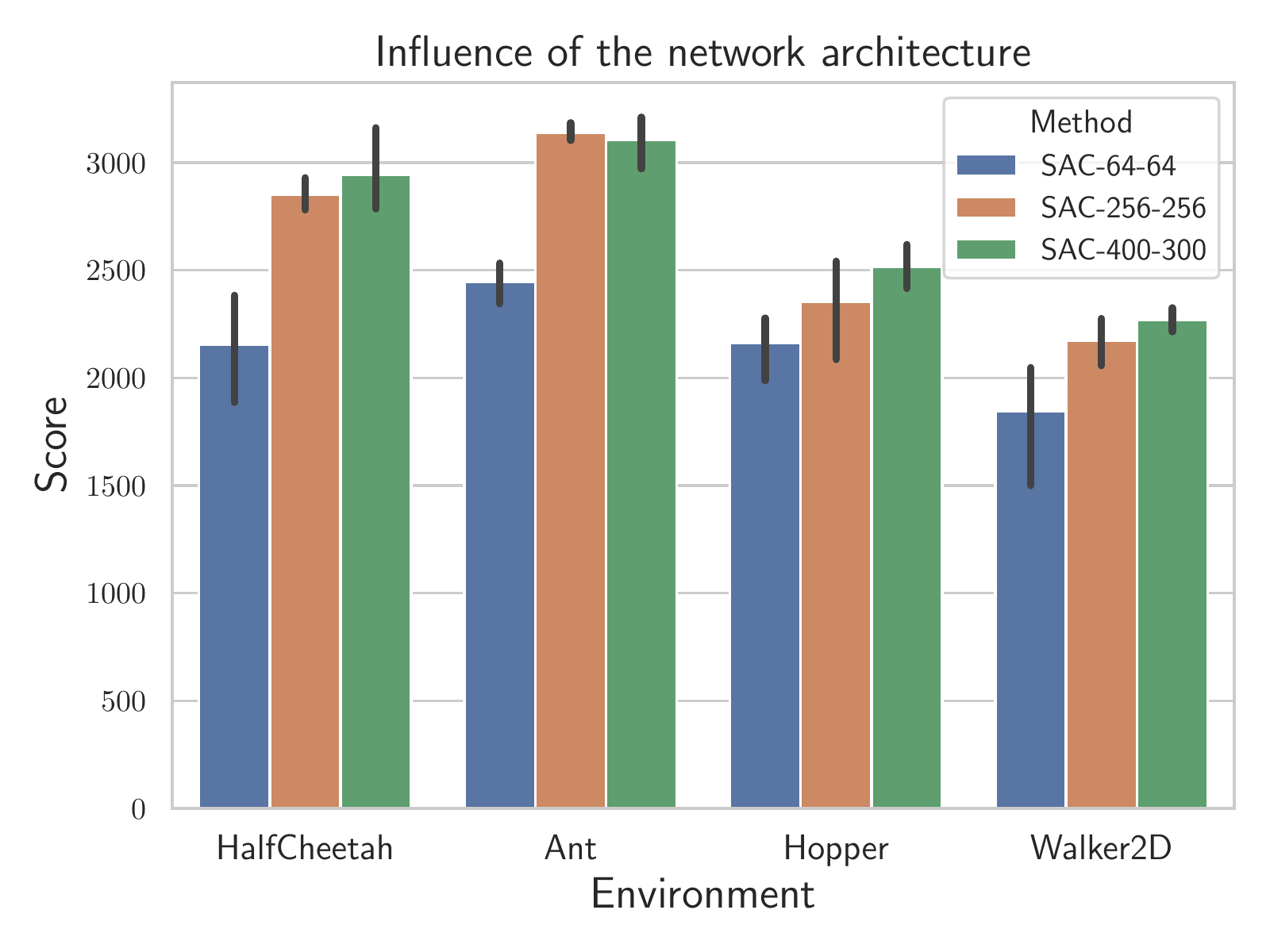}
    \subcaption{Influence of the network architecture}
    \label{fig:net-arch}
  \end{minipage}
  \begin{minipage}[t]{.5\linewidth}
    \centering\includegraphics[width=\linewidth]{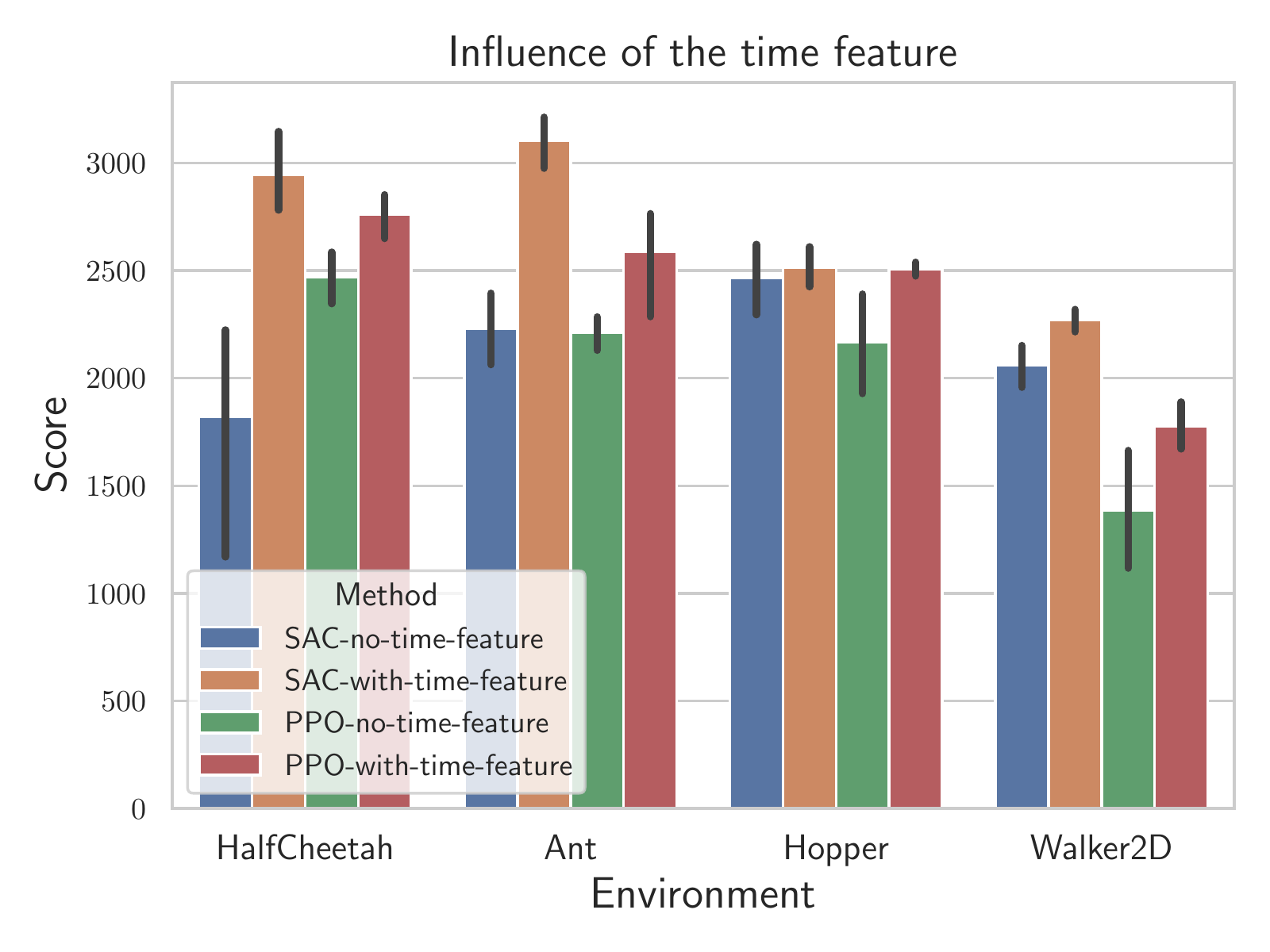}
    \subcaption{Influence of the time feature}
    \label{fig:time-wrapper}
  \end{minipage}
  \caption{(a) Influence of the network architecture (same for actor and critic) for \sac on PyBullet environments. The labels displays the number of units per layer. (b) Influence of including the time or not in the observation for \ppo and \sac.}
\end{figure}

For all experiments with a time limit, as done in \citep{duan2016benchmarking, pardo2017time, rajeswaran2017towards, hill2018stable}, we augment the observation with a time feature (remaining time before the end of an episode) to avoid breaking Markov assumption. This feature has a great impact on performance, as shown in \Cref{fig:time-wrapper}.

~\Cref{fig:net-arch} displays the influence of the network architecture for \sac on PyBullet tasks. A bigger network usually yields better results but the gain is minimal passed a certain complexity (here, a two layers neural network with 256 unit per layer).

\begin{table}[h]
\renewcommand{\arraystretch}{1.1}
\centering
\caption{\sac Hyperparameters}
\label{tab:sac_shared_params}
\vspace{1mm}
  \begin{tabular}{@{}l l| l@{}}
    \toprule
    \multicolumn{2}{l|}{Parameter} &  Value\\
    \midrule
    \multicolumn{2}{l|}{\textit{Shared}}& \\
    & optimizer & Adam~\citep{kingma2014adam}\\
    & learning rate & $7.3 \cdot 10^{-4}$\\
    & learning rate schedule & constant \\
    & discount ($\gamma$) &  0.98\\
    & replay buffer size & $3 \cdot 10^{5}$\\
    & number of hidden layers (all networks) & 2\\
    & number of hidden units per layer & [400, 300]\\
    & number of samples per minibatch & 256\\
    & non-linearity & $ReLU$\\
    & entropy coefficient ($\alpha$) & auto\\
    & target entropy & $- dim(\aspace)$\\
    & target smoothing coefficient ($\tau$)& 0.02\\
    & train frequency & episodic\\
    & warm-up steps & 10 000\\
    & normalization & None\\
    \midrule
    \multicolumn{2}{l|}{\ourSDE}& \\
    & initial $\log \sigma$ & -3\\
    & \ourSDE sample frequency & 8\\
    \bottomrule
  \end{tabular}
\end{table}

\begin{table}[h]
\renewcommand{\arraystretch}{1.1}
\centering
\caption{\sac Environment Specific Parameters}
\label{tab:sac_env_params}
\vspace{1mm}
  \begin{tabular}{@{}l r@{}}
    \toprule
    Environment 	& Learning rate schedule\\
    \midrule
    HopperBulletEnv-v0  & linear\\
    Walker2dBulletEnv-v0 & linear\\
    \bottomrule
  \end{tabular}
\end{table}

\begin{table}[h]
\renewcommand{\arraystretch}{1.1}
\centering
\caption{\tddd Hyperparameters}
\label{tab:td3_shared_params}
\vspace{1mm}
  \begin{tabular}{@{}l l| l@{}}
    \toprule
    \multicolumn{2}{l|}{Parameter} &  Value\\
    \midrule
    \multicolumn{2}{l|}{\textit{Shared}}& \\
    & optimizer & Adam~\citep{kingma2014adam}\\
    & discount ($\gamma$) &  0.98\\
    & replay buffer size & $2 \cdot 10^{5}$\\
    & number of hidden layers (all networks) & 2\\
    & number of hidden units per layer & [400, 300]\\
    & number of samples per minibatch & 100\\
    & non-linearity & $ReLU$\\
    & target smoothing coefficient ($\tau$)& 0.005\\
    & target policy noise & 0.2 \\
    & target noise clip & 0.5 \\
    & policy delay & 2\\
    & warm-up steps & 10 000\\
    & normalization & None\\
    \midrule
    \multicolumn{2}{l|}{\ourSDE}& \\
    & initial $\log \sigma$ & -3.62\\
    & learning rate for \tddd & $6 \cdot 10^{-4}$\\
    & target update interval & 64\\
    & train frequency & 64\\
    & gradient steps & 64\\
    & learning rate for \ourSDE & $1.5 \cdot 10^{-3}$\\
    \midrule
    \multicolumn{2}{l|}{\textit{Unstructured Exploration}}& \\
    & learning rate & $1 \cdot 10^{-3}$\\
    & action noise type & Gaussian\\
    & action noise std & 0.1 \\
    & train frequency & every episode\\
    & gradient steps & every episode\\
    \bottomrule
  \end{tabular}
\end{table}

\begin{table}[h]
\renewcommand{\arraystretch}{1.1}
\centering
\caption{\aac Hyperparameters}
\label{tab:a2c_shared_params}
\vspace{1mm}
  \begin{tabular}{@{}l l| l@{}}
    \toprule
    \multicolumn{2}{l|}{Parameter} &  Value\\
    \midrule
    \multicolumn{2}{l|}{\textit{Shared}}& \\
    & number of workers &  4\\
    & optimizer & RMSprop with $\epsilon = 1 \cdot 10^{-5}$\\
    & discount ($\gamma$) &  0.99\\
    & number of hidden layers (all networks) & 2\\
    & number of hidden units per layer & [64, 64]\\
    & shared network between actor and critic & False\\
    & non-linearity & $Tanh$\\
    & value function coefficient & 0.4\\
    & entropy coefficient & 0.0\\
    & max gradient norm & 0.5\\
    & learning rate schedule & linear \\
    & normalization & observation and reward~\citep{hill2018stable}\\
    \midrule
    \multicolumn{2}{l|}{\ourSDE}& \\
    & number of steps per rollout &  8\\
    & initial $\log \sigma$ & -3.62\\
    & learning rate & $9 \cdot 10^{-4}$\\
    & \textit{GAE} coefficient~\citep{schulman2015high} ($\lambda$) &  0.9\\
    & orthogonal initialization~\citep{engstrom2020implementation} & no \\
    \midrule
    \multicolumn{2}{l|}{\textit{Unstructured Exploration}}& \\
    & number of steps per rollout & 32\\
    & initial $\log \sigma$ & 0.0\\
    & learning rate & $2 \cdot 10^{-3}$\\
    & \textit{GAE} coefficient~\citep{schulman2015high} ($\lambda$) &  1.0\\
    & orthogonal initialization~\citep{engstrom2020implementation} & yes \\

    \bottomrule
  \end{tabular}
\end{table}

\begin{table}[h]
\renewcommand{\arraystretch}{1.1}
\centering
\caption{\ppo Hyperparameters}
\label{tab:ppo_shared_params}
\vspace{1mm}
  \begin{tabular}{@{}l l| l@{}}
    \toprule
    \multicolumn{2}{l|}{Parameter} &  Value\\
    \midrule
    \multicolumn{2}{l|}{\textit{Shared}}& \\
    & optimizer & Adam~\citep{kingma2014adam}\\
    & discount ($\gamma$) &  0.99\\
    & value function coefficient & 0.5\\
    & entropy coefficient & 0.0\\
    & number of hidden layers (all networks) & 2\\
    & shared network between actor and critic & False\\
    & max gradient norm & 0.5\\
    & learning rate schedule & constant \\
    & advantage normalization~\citep{hill2018stable} & True \\
    & clip range value function~\citep{engstrom2020implementation} & no \\
    & normalization & observation and reward~\citep{hill2018stable}\\
    \midrule
    \multicolumn{2}{l|}{\ourSDE}& \\
    & number of workers &  16\\
    & number of steps per rollout &  512\\
    & initial $\log \sigma$ & -2\\
    & \ourSDE sample frequency & 4\\
    & learning rate & $3 \cdot 10^{-5}$\\
    & number of epochs & 20\\
    & number of samples per minibatch & 128\\
    & number of hidden units per layer & [256, 256]\\
    & non-linearity & $ReLU$\\
    & \textit{GAE} coefficient~\citep{schulman2015high} ($\lambda$) &  0.9\\
    & clip range & 0.4 \\
    & orthogonal initialization~\citep{engstrom2020implementation} & no \\
    \midrule
    \multicolumn{2}{l|}{\textit{Unstructured Exploration}}& \\
    & number of workers &  1\\
    & number of steps per rollout & 2048\\
    & initial $\log \sigma$ & 0.0\\
    & learning rate & $2 \cdot 10^{-4}$\\
    & number of epochs & 10\\
    & number of samples per minibatch & 64\\
    & number of hidden units per layer & [64, 64]\\
    & non-linearity & $Tanh$\\
    & \textit{GAE} coefficient~\citep{schulman2015high} ($\lambda$) &  0.95\\
    & clip range & 0.2 \\
    & orthogonal initialization~\citep{engstrom2020implementation} & yes \\

    \bottomrule
  \end{tabular}
\end{table}

\begin{table}[h]
\renewcommand{\arraystretch}{1.1}
\centering
\caption{\ppo Environment Specific Parameters}
\label{tab:env_params}
\vspace{1mm}
  \begin{tabular}{@{}l l r r r@{}}
    \toprule
    & Environment 	& Learning rate schedule & Clip range schedule & initial $\log \sigma$\\
    \midrule
    \multicolumn{2}{l}{\ourSDE}& \\
      & AntBulletEnv-v0 & default & default & -1\\
      & HopperBulletEnv-v0 & default & linear & -1\\
      & Walker2dBulletEnv-v0 & default & linear & default\\
    \midrule
    \multicolumn{2}{l}{\textit{Unstructured Exploration}}& \\
      & Walker2dBulletEnv-v0 & linear & default & default\\
    \bottomrule
  \end{tabular}
\end{table}

\end{document}